# Creating *emoji* lexica from unsupervised sentiment analysis of their descriptions

Milagros Fernández-Gavilanes[a,*], Jonathan Juncal-Martínez[a], Silvia García-Méndez[a], Enrique Costa-Montenegro[a], Francisco Javier González-Castaño[a]

[a] *GTI Research Group, Telematic Engineering Department, School of Telecommunication Engineering, University of Vigo, Vigo, Pontevedra 36310, Spain*

**Abstract**

Online media, such as blogs and social networking sites, generate massive volumes of unstructured data of great interest to analyze the opinions and sentiments of individuals and organizations. Novel approaches beyond *Natural Language Processing* are necessary to quantify these opinions with polarity metrics. So far, the sentiment expressed by *emojis* has received little attention. The use of symbols, however, has boomed in the past four years. About twenty billion are typed in Twitter nowadays, and new *emojis* keep appearing in each new Unicode version, making them increasingly relevant to sentiment analysis tasks. This has motivated us to propose a novel approach to predict the sentiments expressed by *emojis* in online textual messages, such as tweets, that does not require human effort to manually annotate data and saves valuable time for other analysis tasks. For this purpose, we automatically constructed a novel *emoji sentiment lexicon* using an unsupervised sentiment analysis system based on the definitions given by *emoji* creators in `Emojipedia`. Additionally, we automatically created lexicon variants by also considering the sentiment distribution of the informal texts accompanying *emojis*. All these lexica are evaluated and compared regarding the improvement obtained by including them in sentiment analysis of the annotated datasets provided by Kralj Novak et al. (2015). The results confirm the competitiveness of our approach.

## 1. Introduction

*Emojis* are commonly used in smartphone texting, social media sharing, advertising, and more. For example, in 2015 nearly half of all texts posted on Instagram contained them (Dimson, 2015). Similarly, at the time of this research, in a 1% random sample of tweets published from July 2013 to August 2017, 19.88 billion tweets contained *emojis* according to `Emojitracker.com`[1]. *Emojis* differ from emoticons in that the former are represented by pictographs with a designated textual description, while the latter are typographic facial representations.

Even though *emojis* seem a recent alternative to emoticons, they have been around for 30 years. They were first used in Japan (*emoji* literally means "*image*" and "*character*") and originally could only be used on Japanese phones (D'Aleo et al., 2015). They gained popularity when the Unicode standard incorporated them and Apple included them in its operating systems in 2011.

*Corresponding author: Tel: +34 986 814081
*Email addresses:* `milagros.fernandez@gti.uvigo.es` (Milagros Fernández-Gavilanes), `jonijm@gti.uvigo.es` (Jonathan Juncal-Martínez), `sgarcia@gti.uvigo.es` (Silvia García-Méndez), `kike@gti.uvigo.es` (Enrique Costa-Montenegro), `javier@det.uvigo.es` (Francisco Javier González-Castaño)

[1] `http://www.emojitracker.com/api/stats`

Since then, their number has continuously grown with the introduction of new characters in each new Unicode version, including not only faces but also pictographs representing concepts and ideas such as weather, vehicles and buildings, food and drinks, animals and plants, and emotions, feelings or activities, like running and dancing (Pavalanathan & Eisenstein, 2015).

Moreover, in 2007, Google completed the conversion of "*enhanced emotions*" to Unicode private-use codes, and in 2009 a set of 722 Unicode characters was defined collecting all Japanese *emoji* characters. More pictographs were added in 2010, 2012 and 2014 (Davis & Edberg, 2017). In November 2013, a study indicated that 74% of the United States population used these graphic symbols[2]. In China, the percentage of population that used them in nonverbal communications was even higher, reaching 82% (Statista, 2013; Sternbergh, 2014).

This suggests the capability of *emojis* to express feelings or emotions in absence or other elements such as words, facial expressions or voice cues (Wallbott & Scherer, 1986), across different cultures (D'Aleo et al., 2015), which means that they can be exploited as *a priori* knowledge about opinions in user comments (Hu et al., 2013). Consequently, they are appealing to *Sentiment Analysis* (SA), a subfield of *Natural Language Processing* (NLP). The latter combines computational science methods (such as artificial intelligence, automatic learning, or statistical inference) with applied linguistics to achieve computer-aided comprehension and processing of information expressed in human language. In this scenario, SA, also called *opinion mining*, is the field that analyzes people's opinions, sentiments, evaluations, appraisals, attitudes, and emotions towards entities such as products, services, organizations, events and topics, and their attributes (Liu, 2012). Although linguistics and NLP have a long history in common, little research about people's opinions and sentiments was conducted before 2000. Since then, it has become a very active research area, especially in the analysis of informal texts such as tweets. Only in recent years emoticons have been considered to play a role (Boia et al., 2013; Davidov et al., 2010; Hogenboom et al., 2015; Solakidis et al., 2014; Yamamoto et al., 2014), albeit nowadays *emojis* are more popular. Even so there is still little research work devoted in both of them (Guibon et al., 2016).

One of the most evident issues is the disparity of appearance of an *emoji* from one platform to another. Figure 1 shows the *emojis* corresponding to *grinning* 😀 and *crying* 😢 for two different platforms, Apple and Google. It is considered that any symbolic representation based on a given name is totally valid, although the meaning associated to each symbol is unique: for example, the *emoji grinning face* 😀 refers to a positive emotion, whereas the *emoji crying face* 😢 clearly has a negative meaning.

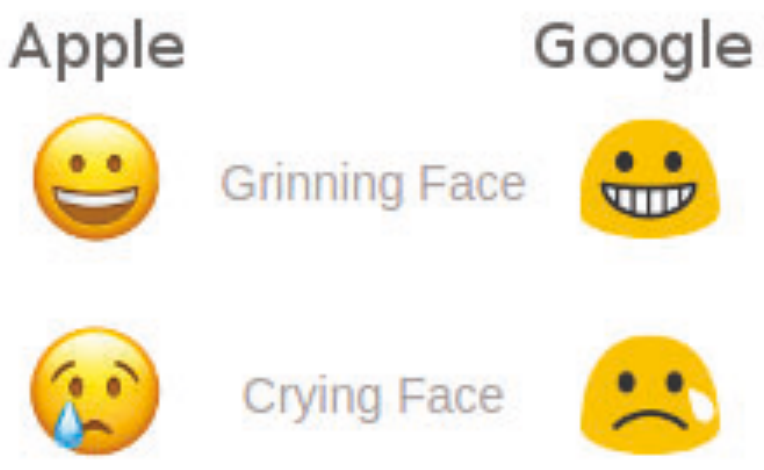

Figure 1: Example of positive and negative emojis.

Due to the variability of *emoji* representations in different platforms, and given the continuous introduction of new *emojis* in each new Unicode version, it is very difficult to understand meanings beyond affective stances in terms of positivity, neutrality or negativity, and those can vary with social context and author identity (Derks et al., 2007; Park et al., 2013; Schnoebelen, 2012). That is, in some cases the original meaning has nothing in common with that attributed by people in a particular context, and could be quite different to the initial intention of the creator.

[2] https://blog.swiftkey.com/the-united-states-of-emoji-which-state-does-your-emoji-use-most-resemble/.

Accordingly, some authors have constructed *emoji* sentiment lexica by manually annotating on informal texts (with the consequent arduous work that this entails) (Kralj Novak et al., 2015). Therefore, it is interesting to adopt an approach that initially considers *emoji* information that does not require human annotation, such as the real meaning of an *emoji*, which is given by its definition, which, in turn, is still strongly linked to the emotional meaning conceived by its creators.

In this paper we present our research to *automatically* construct sentiment lexica with 840 *emojis* using an *Unsupervised System with Sentiment Propagation Across Dependencies* (USSPAD) approach, based on the analysis of the sentiment of informal texts in English and Spanish. The initial sentiment of each *emoji* is derived from a sentiment score obtained after applying the meaning assigned by its creator. Then this value is improved taking into account sentiment scores obtained from informal texts in which that *emoji* appears. So, the results reflect not only the actual use of *emojis* in a context, by applying SA to informal texts such as Twitter, but also the sentiments in the definitions describing such *emojis* in `Emojipedia`[3]. To the best of our knowledge, this is the first time that *emoji* definitions are considered in automatic *emoji sentiment lexicon* creation, where textual information is analyzed with USSPAD, and later combined with textual contexts.

Different experiments and results are presented. In this regard, comparing different approaches is extremely difficult due to the lack of a goldstandard *emoji sentiment lexicon*. Consequently, we compare our strategies with the few in which *emojis* were subject to SA, providing support for our main hypothesis. As a testbed, we employed the available annotated datasets provided by Kralj Novak et al. (2015). Only considering the "initial" sentiment of the *emojis* (i.e. by only taking their short names into account), our approach was competitive with that of Kralj Novak et al. (2015) (based on annotated data), and significantly better when also considering their definitions and usage contexts (messages contexts then are included in). Note that, unlike that approach, ours is fully unsupervised. At the same time these results confirm that *emoji* descriptions add discriminating information that could be exploited in more advanced social NLP systems, given the improvement in accuracy and macroaveraging metrics they achieve.

The paper is organized as follows. Section 2 reviews related work on *emoji* SA. Section 3 describes the proposed SA system. Section 4 discusses experimental results for Twitter dataset. Finally, Section 5 summarizes the main findings and conclusions.

## 2. Related work

In spite of the fact that *emojis* may be considered a language form, they have been little studied from an NLP perspective, in contrast to their predecessors the *emoticons*. The few exceptions include studies on *emojis* usage and semantics.

For example, Barbieri et al. (2016a) constructed a vector space model aiming at providing a common semantic ground in which *emojis* are naturally distributed according to geolocation in metropolitan areas. In (Barbieri et al., 2016b), the study was extended to countries with different languages. Finally, Ljubešić & Fišer (2016) investigated the global distribution of *emojis*, performing a cluster analysis over countries and a correlation analysis between *emoji* distributions and World Development Indicators.

Regarding semantics studies, Barbieri et al. (2016c) generated, validated, and shared semantic vectorial models built over 10 million tweets posted by USA users by consistently mapping in the same vectorial space both words and *emojis*. They applied skip-gram word embedding models (effectiveness was validated by comparing the output of these models with human assessment using semantic similarity experiments). Their aim was to estimate the degree of similarity between two *emojis* in a situation where both can occur. Later, Eisner et al. (2016) used a similar distributional semantic models, but instead of running skip-gram models on large collections of *emojis* and their tweet contexts, *emoji* embeddings were directly trained on Unicode short *Common Locale Data Repository (*CLDR*)* names[4] (thus requiring much less training data).

Our research is more focused on acquisition of the *emojis'* emotional meaning (in the form of polarity values) by obtaining a lexicon from their descriptions. In our work, SA allows the polarity of a text to be determined (positive, neutral, negative, or any degrees of these polarities). The current surge of research interest in this kind of systems is attributed to the fact that explicit information on user opinions is often hard to find, confusing, or overwhelming (Hogenboom et al., 2013). Many approaches exist where SA is generally accomplished by: (i) unsupervised systems combined with a sentiment lexicon (Kolchyna et al., 2015); (ii) simple machine learning algorithms (Moraes et al., 2013; Narayanan et al., 2013); and/or (iii) more advanced ones using deep learning (dos Santos & Gatti, 2014). Harvesting information from *emojis* has rarely been explored. Yet nowadays a typical first step when preprocessing a text is to remove many of the Unicode symbols, which often reflect *emojis*. Therefore, their contributions to the sentiment of the text is lost (Tauch & Kanjo, 2016). An *emoji* often carries the whole emotional weight. For example, in the text *"I still have Christmas shopping to do 😓"*, none of the components of the sentence (except the *emoji* itself) contribute to the sentiment of the expression. But, if the *emoji* is taken into account, it contributes to a negative emotion, and the overall expression acquires a negative sentiment. In this respect, there are few different studies on how to capture the real emotion expressed by an *emoji*. We can distinguish between two strategies: manual, and semiautomatically or automatically labeling.

[3] `http://emojipedia.org/`

[4] These are annotations which provide names and keywords for Unicode *emojis*, which are available at http://unicode.org/emoji/charts/emoji-list.html .

Manual labeling has been used in preliminary studies centered on emoticons, which were manually rated with positive or negative sentiments (Hogenboom et al., 2013) or with other sentiment annotations such as extremely negative, extremely positive, and neutral (Agarwal et al., 2011). Another example is (Miller et al., 2016), who surveyed the interpretation of a sample of the most popular *emoji* Unicode characters (available on platforms like Apple, Google, Microsoft, Samsung and LG) with the help of human annotators. They tried to measure the variance of sentiment (positivity, neutrality, or negativity) and semantics (meaning). The results showed that the variance of interpretation was only low for 4.5% of the *emojis*, whereas for 25% of them the participants disagreed. In addition, when considering representations accross different platforms, the discrepancy increased. Their main conclusion was that *emojis* may lead to misunderstandings.

In Kralj Novak et al. (2015), the sentiments of *emojis* were calculated from the sentiment of tweets. In this research, 83 native speakers of different languages labeled a large number of tweets manually as positive, neutral, or negative. The occurrences of each *emoji* were counted, with the tweet labels containing it determine its sentiment, and its discrete probability distribution was calculated. Finally, in Er et al. (2016), the authors selected 78 strong subjetive *emojis* and 34 weak subjective from a list of 751 *emoji* characters obtained by Kralj Novak et al. (2015). Subsequently, they assigned scores of +2 and -2 to strong positive and strong negative *emojis*, respectively, and +1 and -1 to weak positive and weak negative *emojis*, respectively.

Very few studies have tried to assign polarities to *emojis* semiautomatically or automatically. For example, Hussien et al. (2016) collected 134,194 Arabic tweets and analyzed which *emojis* were most frequently used in them. The most frequently used ones were classified into four categories: anger, disgust, joy, and sadness. Finally, a weight between -5 and +5 was assigned to each *emoji*, where the signs - and + represent negativity or positivity of the sentiment, depending on the category. This weight was obtained from the *Årup Finn (AFINN)* lexicon (Nielsen, 2011), which contains some *emojis* as entries. For those *emojis* not included, the authors took their CLDR short names, performed a search for its words in the commented lexicon (independently, i.e. word by word) and manually assigned the weights from the category to which the *emojis* belonged.

We are only aware of other two works with fully unsupervised approaches: Lu et al. (2016) and Kimura & Katsurai (2017). In the first work, Lu et al. (2016) analyzed *emoji* use in text messages by country, concluding that the most employed ones express some emotion. Then, each *emoji* was assigned its official CLDR short name as its translation to text. To automatically obtain the associated sentiment, each short name was processed with the LIWC tool[5] (processing the words in the short name independently without taking into account their dependencies). In total, the authors obtained sentiment scores for 199 *emojis* in an automatic way. Note, however, that they neither exploited the *emojis* definitions nor their usage contexts. In section 4, we will show that, depending on the language of the dataset used and the comparison metric, our approach outperforms theirs by 3%-8%.

[5]Available at https://liwc.wpengine.com/. In summary, this tool counts the words in a text that express different emotions.

In the second work, Kimura & Katsurai (2017) first extracted, for each word of a tweet [6] that co-occur with a target *emoji*, the set of *Wordnet* synsets to which that word belongs, and then they retrieved the most frequent affective label from *WordNet-Affect* (Strapparava & Valitutti, 2004). They distinguished between 5 sentiment categories, *happiness*, *disgust*, *sadness*, *anger* and *fear*, constructed following a hierarchical structure. Once they obtained the labels for all the words that co-occur with an *emoji*, they calculated an *emoji* sentiment score vector based on the co-occurrences between *emoji* and sentiment words. Finally, the *emoji* score resulted from substracting all the negative scores from the positive ones. In total, the authors obtained sentiment scores for 236 *emojis* automatically. Note that, just as in (Lu et al., 2016), they neither exploited *emoji* definitions nor their usage contexts. In section 4 we report improvements of our approach over theirs of 4%-6% for most datasets under study.

The main objective of our research is to provide an *emoji sentiment lexicon*, in other words, to automatically predict whether an *emoji* expresses positive, negative, or neutral sentiments without the need for supervision. The review of the state-of-the-art reveals that most existing approaches assign sentiment scores to *emojis* manually (Er et al., 2016; Miller et al., 2016), simply applying to the *emojis* the sentiment of the texts where they are included (Kralj Novak et al., 2015). They considered the words included in short Unicode names independently (Hussien et al., 2016; Lu et al., 2016) or affective information present in tweets (Kimura & Katsurai, 2017), without taking into account linguistic aspects such as negations (affective labels differ completely if extracted from afirmative or negative sentences).

In our case, lexica are automatically constructed using an USSPAD, not only taking into account the sentiment score of the informal texts where *emojis* appear, but also considering the emotion expressed by their creators in their descriptions. The unsupervised system not only automates the process of obtaining the lexicon (without the need for any type of training), but also provides competitive results. Unlike previous studies which only considered isolated words, our approach tries to comprehend the semantics of linguistic constructions, such as negation.

## 3. System overview

Although machine learning algorithms (including complex neural networks) have proven to be extremely useful in the field of SA, an obvious disadvantage is that they are not immediately applicable to domains other than the domain they were designed for, unless re-adapted with additional techniques such as transfer learning (Calais Guerra et al., 2011; Yoshida et al., 2011; Medhat et al., 2014). Moreover, classifier training requires labeled datasets (Moreno Ortiz & Pérez Hernández, 2013), which are often difficult or even impossi-ble to obtain. This is mainly because labeling data takes considerable human effort. Our USSPAD overcomes this disadvantage (Fernández-Gavilanes et al., 2015).

It is very important to identify the different steps to be applied to obtain an initial lexicon only from the meaning assigned by the *emoji* creators. Our initial methodology is shown in Figure 2: (1) acquisition of both informal texts with *emojis* and the *emoji* definitions from `Emojipedia` (Section 3.1); (2) SA of descriptions and informal texts through propagation using an USSPAD with a sentiment lexicon (Section 3.2); (3) creation of *emoji* sentiment lexicon (Section 3.3). Summing up, *emojis* are extracted from a set of informal texts and their descriptions are acquired from the `Emojipedia` repository. NLP techniques then capture their linguistic peculiarities, and an *emoji* sentiment lexicon is created in order to be later used to improve sentiment detection performance for informal texts subject to a similar process of SA.

In Figure 2, the green arrow refers to gathering a set of informal texts with *emojis*, whereas the red and blue arrows refer to the process carried out individually on each *emoji* description (from extraction from `Emojipedia` until the estimation of its sentiment value and its inclusion in the *emoji sentiment lexicon from descriptions*), as well as on particular informal texts to obtain their sentiment values, considering the polarity of *emojis* previously analyzed. Below we explain the process in greater detail.

[6] They collected a Twitter dataset with 414.977 tweets.

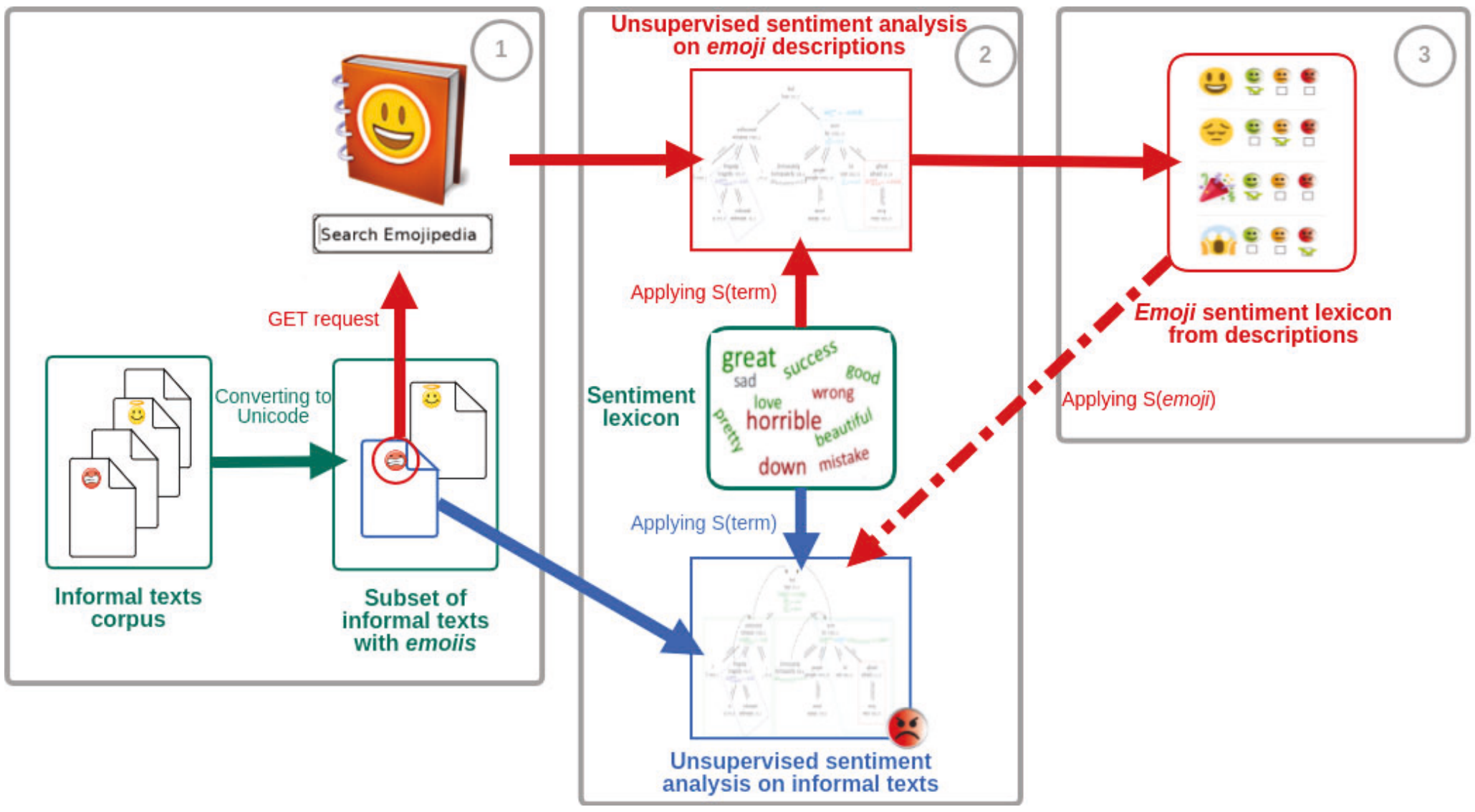

Figure 2: Methodology to produce an *emoji sentiment lexicon* from *emoji* descriptions.

### *3.1. Acquiring informal texts with emojis and emoji definitions from* `Emojipedia`

After gathering a large corpus of informal texts, we selected a subset with *emojis*. As we are focusing on the writing styles of individual users, we eliminate spam and automated texts belonging, for example, to organizations (Yardi et al., 2010). Moreover, repeated content also has to be removed, such as Twitter messages with the "`RT`" token.

In order to extract *emoji* characters from informal texts, the messages must be converted into a Unicode representation and regular expressions must be used for the extraction[7]. Unicode is a text encoding schema that translates characters from almost all extant languages into bytes[8]. It provides a complete list of *emoji* characters, included in the *Emoji & Pictographs* category (other categories include non-Roman characters such as different numbering systems and mathematical symbols), with images from different vendors, version and source information, default style, and annotations like manually curated keywords.

Once a subset of informal texts with *emojis* is retrieved, the next step is to get their descriptions from the `Emojipedia` resource. This is a sort of *emoji* dictionary in English, that includes images from a variety of platforms. It allows querying through GET requests of codes in *UTF-8 Hex Bytes*. Accordingly, each *emoji* Unicode codepoint in hexadecimal notation is converted to *UTF-8 Hex Bytes* and then submitted by a GET request to http://emojipedia.org/search/?q=. For example, the *emoji face with tears of joy* 😂, with Unicode "`0x1f602`", after conversion to *UTF-8 Hex Bytes*, is "`F0 9F 98 82`". Thus, the query would be: http://emojipedia.org/ search/?q=%F0%9F% 98%82. Once the response to that request has been obtained (in HTML format), it is parsed through *jsoup* (Hedley, 2016).

### *3.2. Sentiment analysis of* emoji *descriptions and informal texts*

At this point, and in order to predict whether an *emoji* expresses a positive, negative, or neutral sentiment without human supervision, our approach performs SA on both the informal texts containing the *emojis* and their definitions.

Most learning- or lexicon-based state-of-the-art systems only take into account isolated words and not the relationships between them (Hu & Liu, 2004; Pak & Paroubek, 2010; Turney, 2002). Some try to simulate comprehension of some linguistic constructions, such as negation, but this fails often due to the complexity of human language (König & Brill, 2006; Pang & Lee, 2008; Quinn et al., 2010). Other systems analyze dependencies between words but rely on supervised systems to interpret human language (Nakagawa et al., 2010; Zhang & Singh, 2014). Our approach makes use of an unsupervised method to exploit the linguistic information present in dependencies retrieved from a parsing analysis. NLP techniques capture linguistic peculiarities that improve sentiment detection performance. Our approach has four steps: (3.2.1) preprocessing, (3.2.2) lexical and syntactic analysis, (3.2.3) application of a sentiment lexicon and (3.2.4) sentiment analysis through propagation (Fernández-Gavilanes et al., 2016).

[7]This process was carried out using the `Emoji-java` library, available at `https://github.com/vdurmont/emoji-java`.
[8]See http://www.unicode.org

#### *3.2.1. Preprocessing informal texts and descriptions*

The unsupervised approach applied in this research was initially developed for SA of online texts, which pose several challenges for NLP. The language used in social media is quite different from that employed in other fora because it often contains words that cannot be found in dictionaries, often used with particular orthographic and typographical characteristics. This is the reason why a previous preprocessing step is required. It restores as much as possible the message language to natural language by eliminating atypical expressions and so minimize noise in later stages (Fernández-Gavilanes et al., 2015; Fernández-Gavilanes et al., 2016). Examples are replacement of hashtags (such as "*#hashtag*"), usernames (such as "*@username*") and URLs (such as "*http://url*"), as well as replacement of repeated characters in a word to reduce it to its base form (e.g. "*coooooool*" is replaced by "*cool*"), and replacement of abbreviations by their expresions (such as "*LOL*" by "*laughing out loud*").

In the case of *emoji* descriptions, the language used is more formal. For this reason, other treatments were applied after studying their content. The procedure is similar to that for informal texts. However, there are some peculiarities that deserve mention. Some descriptions may in turn include more *emojis*. For example, the *emoji face with tears of joy* 😂 starts with the following sentence in `Emojipedia`: "*A laughing emoji which at small sizes is often mistaken for being tears_of_sadness* 😭 ". The underlined text is the Unicode name of a new *emoji*. In this case, the Unicode name is replaced by its corresponding Unicode codepoint, marked by a pair of tags: "`[emoji]`" to open, and "`[/emoji]`" to close.

Both descriptions and informal contents may include named entities that need to be recognized. *Named-entity recognition* (NER) is a relevant NLP problem, which is very useful in information extraction. It seeks to locate, and classify entities into one of the predefined labels, such as person name, organization name, location name, and dates (Das et al., 2017). In this step, our approach applies the *Stanford Named Entity Recognizer* (Finkel et al., 2005) library. For example, the description of the *emoji face with ok gesture* 🙆 retrieved from `Emojipedia` contains the following text: "*A person with arms above his or her head ...but shown in the Apple artwork as a woman, and by Microsoft as a man...in 2010 ...*". In the first case, "*Apple*" and "*Microsoft*" were detected and replaced by "*Organization*", whereas "*2010*" was replaced by "*Date*".

#### *3.2.2. Lexical and syntactic analysis of descriptions and informal texts*

In order to derive the syntactic context, following the procedure carried out in (Fernández-Gavilanes et al., 2016), each preprocessed *emoji* description or informal text is split into tokens and then into sentences. To ensure that all inflected forms of a word are covered, this stage performs lemmatization and *part-of-speech* (POS) tagging using the *FreeLing Tagger* (Atserias et al., 2006; Padró & Stanilovsky, 2012), or, more specifically, its tagger implementation based on hidden Markov models (HMM) (Brants, 2000). *FreeLing* is a library that provides multiple language analysis services for English and Spanish, including probabilistic prediction of categories for unknown words. POS tagging allows lexical items to be identified, such as adjectives, adverbs, verbs and nouns, that contributes to the correct recognition of sentiment in texts. The accuracy rate is around 90% for formal texts, but often lower in the case of informal texts (social media, blogs, and fora). The resulting lemmatized and POS-annotated *emoji* descriptions or informal texts feed a parser that transforms the output of the tagger into a full parse tree. Finally, the tree is converted to dependencies, and the functions are annotated with the *FreeLing Parser* (Padró & Stanilovsky, 2012).

For example, returning to the description of the *face with tears of joy* 😂 *emoji*, the preprocessing step returns the following: "*A laughing emoji which at small sizes is often mistaken for being [emoji]U+1F62D*

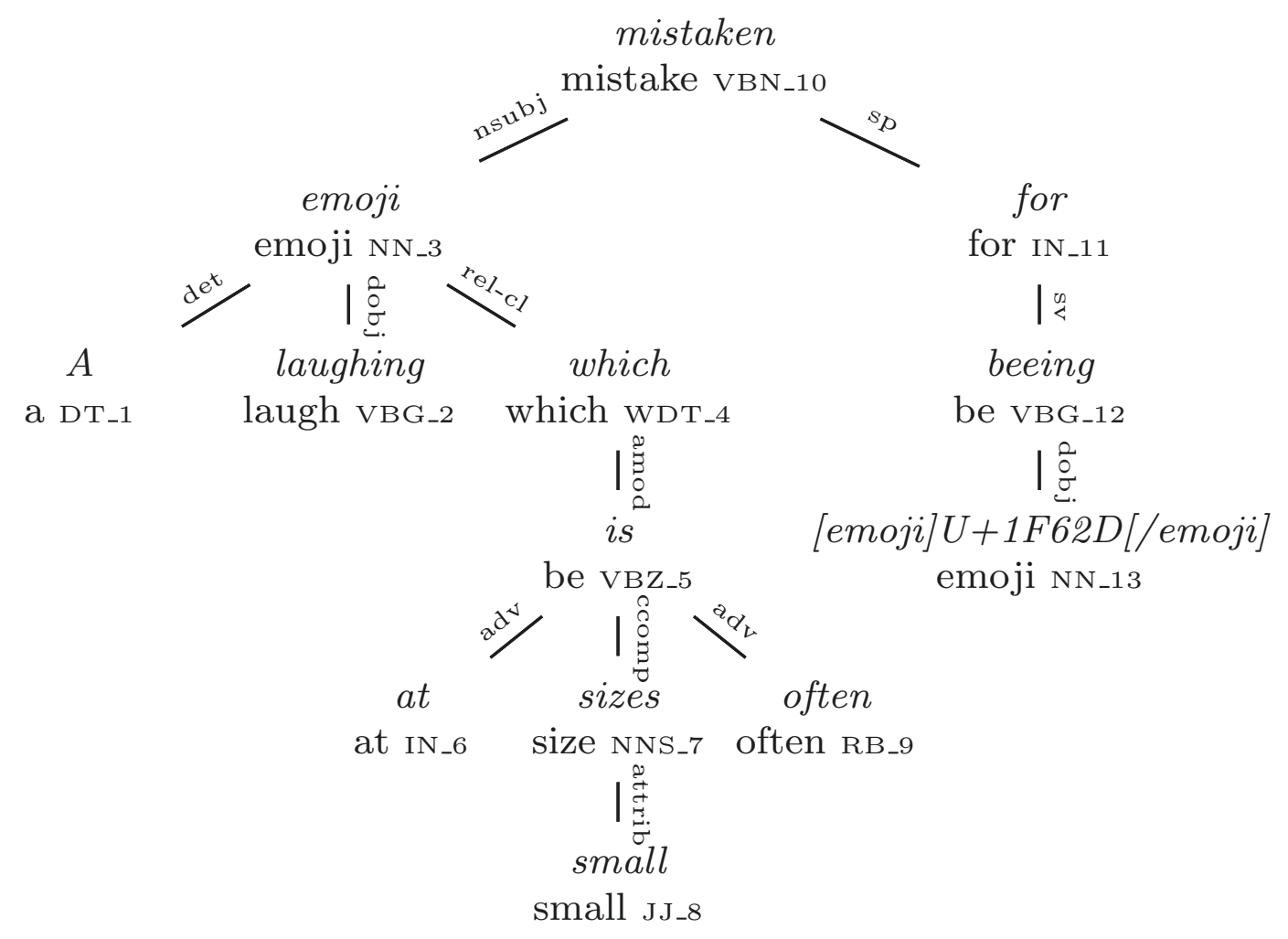

Figure 3: Dependency parse for the running example.

*[/emoji]*”. The final result of applying the tagger and the dependency parser for English is a tree structure with nodes and edges, where each node includes lexical information such as form, lemma, and tag, such as in Figure 3. In addition, each node will be related to other nodes on which it depends syntactically, connected by edges. For example, the token “*laughing*” is a gerund located at the second position of the sentence, represented as *VBG_2*, whose lemma is *“laugh”*. This node only depends syntactically on one other node, “*emoji*”, with the same lemma and represented by *NN_3*, which implies that there is a common noun in the third position.

*3.2.3. Sentiment lexicon*

Sentiment lexica (also called *polarity* or *opinion lexica*) have been used in many systems to help determine the semantic orientation of the sentences contained in a text. For example, it has been shown that it is possible to aggregate polarity values from a sentence or a document and compute the resulting sentiment on a continuous scale (Fahrni & Klenner, 2008; Missen & Boughanem, 2009; Tsytsarau et al., 2010). More sophisticated methods employ strategies involving lexis, syntax and semantics (Quinn et al., 2010). Sentiment lexica are usually lists of words with associated values representing their sentiment, usually integers expressing polarity and polarity intensity as increasing or decreasing absolute values. There exist multiple sentiment lexica, created in different ways: (i) manually, the most complex and time consuming to obtain; (ii) automatic; and (iii) semiautomatic. The literature describes approaches to the last two types. Turney & Littman (2002) estimated the correlation of a word with other positive and negative predefined words. Hu & Liu (2004) presented another dictionary-based approach, where a group of initial words or seeds was expanded from synonym and antonym relationships given by dictionaries like WordNet. For example, in the case of SentiWordNet (Esuli & Sebastiani, 2006) the words contained in the same categories of WordNet will have similar polarities.

In our case, the informal texts and descriptions associated with each *emoji* are generic (independent of the context). Due to this fact and from our results in Fernández-Gavilanes et al. (2015) and Juncal-Martínez et al. (2016) (specifically, in SemEval tasks 10 and 4 of SA in Twitter, respectively), we decided to use the *Semantic Orientation CALculator* (SO-CAL) dictionaries (Taboada et al., 2011) and AFINN (Nielsen, 2011) general purpose sentiment lexica. Since there are words in SO-CAL that are not contained in AFINN and vice versa (we consider these words “uncommon”), as well as words with slightly different polarity values in them, we merged both lexica to obtain a unique list with the corresponding polarity values. Uncommon words were entered directly into the new list, with values between -5 (the most negative polarity value) and 5 (the most positive polarity value) as a shared scale for both lexica. In the case of common words (those in both lexica), we performed an unweighted average of the polarity values assigned to a given word in each independent lexicon, to obtain a new polarity value, again between -5 and 5. The resulting new sentiment lexicon was applied in this research.

#### *3.2.4. Sentiment analysis through propagation*

Our unsupervised approach for SA, USSPAD, described in (Fernández-Gavilanes et al., 2016), is based on the concept of *sentiment propagation* in sentence NLP. After the syntactic analysis of the sentence, and given the lexical polarities of the words included in its dependency parse as real numbers (from a sentiment lexicon), we propagate these real polarity scores across the dependency parse. Throughout this propagation we deal with different linguistic structures that transform sentiment, to finally assign an overall positive, negative or neutral polarity to the sentence. The corresponding algorithms are described in detail in (Fernández-Gavilanes et al., 2016). Basically, they are:

- Intensification treatment

  Intensifiers and diminishers are linguistic terms (usually adverbs and some adjectives) that refer to elements that do not contribute to the propositional meaning of a sentence but emphasize or attenuate the semantic orientation of the words or expressions that they accompany. A list of particles provides a scale of strength values beyond their negative or positive orientation (Brooke, 2009).

  For example, the description of the *emoji pouting cat face* 😾 contains the expression *"very grumpy"*. The word "*grumpy*" expresses by itself negativity, while the word "*very*" is an intensifier that emphasizes the strength of a word or expression and provides an emotional context. For this reason, the expression "*very grumpy*" is more negative than "*grumpy*" alone.

- Modification treatment

  In many cases, the meaning of a word can be altered when accompanied by another word that changes it. The latter can add sentiment to the former (Klenner et al., 2014; Tron, 2013).

  For example, the description of the *emoji two hearts* 💕 includes an expression referred to "*love hearts*", where both terms express positive sentiment, but the first accentuates the second.

  Another example is, in the description of the *emoji black heart* 🖤, the expression "*dark humor*". These words have totally opposite polarities: "*humor*" is a word that expresses positivity, while "*dark*" has a negative connotation. However, their union would be associated with a negative context.

- Negation treatment

  Negation is present in all human languages and it is used to reverse the polarity of parts of a statement. Besides their direct meaning, negated statements often carry a latent positive meaning.

  Different approaches treat this linguistic structure in SA (Mejova & Srinivasan, 2011; Pang et al., 2002; Polanyi & Zaenen, 2004). In our approach the first step is detecting negator terms from a list created from different previous works (Carrillo de Albornoz & Plaza, 2013; Councill et al., 2010; Zhang et al., 2012), indicating where the negation scope begins. The next step is to locate the end of the scope using the syntactic procedure explained in (Fernández-Gavilanes et al., 2015). Once the full negation scope is detected, its polarity is calculated.

  For example, the description of the *emoji unamused face* 😒 contains the sentence "*This face is not amused*". In this case, the word "*not*" is a negator node that implies a negative context. The negation scope includes the node "*amused*" with positive polarity, so that the apparent positivity of the sentence turns into negativity.

- Adversative/concessive treatment

  Adversative and concessive clauses are constructions that express antithetical circumstances (Crystal, 2011). In both cases, one part of the sentence is in contrast with the other, but the kind of connector (Allan & Brown, 2010) determines which part of the construction is more important. For instance, in adversative sentences, it is considered that the part containing the connector is the most important (Poria et al., 2014), while in concessive sentences, it is the least important (Rudolph, 1996).

For example, in the *emoji* description *relieved face* 😌, in the sentence "*Happy, but not over the top*", the USSPAD in our approach detects the adversative clause through connector "*but*". The sentiment associated with the adversative clause is amplified by a certain value (as if it was an intensification), while the other clause is attenuated. In the end, the outcome is the sum of both sentiment values (Fernández-Gavilanes et al., 2016).

### *3.3. Emoji sentiment lexicon generation*

Once all the previous steps have been performed on the descriptions of the *emojis* included in the subset of informal texts, the *emoji* sentiment lexicon can be created. At this point, our USSPAD produces a polarity score for each *emoji* from the descriptions. This valuable information can then be used as another sentiment lexicon to improve the SA of the informal texts.

## 4. Evaluation and experimental results

In this section we evaluate the proposed architecture after obtaining the *emoji sentiment lexicon from descriptions*. Our objective is to determine if the definitions of *emojis* are useful by themselves or not. At the same time, even though *emojis* are ubiquitously used as a simple language (Lu et al., 2016), we are interested in assessing if users from different languages behave similarly when using them and, simultaneously, whether they interpret their meaning in a similar way. Additionally, we study how to adapt our architecture to automatically create lexicon variants in combination with the sentiment distribution of the informal texts, where *emojis* are embedded for the different languages used in a dataset.

### *4.1. Datasets*

We tested the performance of our methodology on publicly available annotated datasets with over 1.6 million annotated tweets (Kralj Novak et al., 2015; Mozetič et al., 2016). The languages covered are: Albanian, Bulgarian, English, German, Hungarian, Polish, Portuguese, Russian, Ser/Cro/Bos (a joint set of Serbian, Croatian, and Bosnian tweets, which are difficult to distinguish in Twitter), Slovak, Slovenian, Spanish, and Swedish. These data are available as 15 language files in `csv` format from the public language resource repository CLARIN[9]. For each language and for each labeled tweet there is a tweet ID (as provided and required by Twitter), a sentiment label (negative, neutral, or positive), and an annotator ID (anonymized).

These tweets were posted between April 2013 and February 2015. All tweets, except the English ones, were collected during a joint project of *Gama System*[10] using a platform called *PerceptionAnalytics*[11], and acquired through *Twitter Search API*, with geolocations of the largest cities. For English tweets, the mechanism employed was to retrieve a random sample of 1% of all public tweets through *Twitter Streaming API*, and then to filter out the English posts (Mozetič et al., 2016). In our case, we focus on the tweets containing *emojis* in two datasets:

- The English dataset, a collection acquired in September 2014, composed of 103,034 tweets (26,674 negative, 46,972 neutral, and 29,388 positive tweets), manually labeled by 9 annotators, of which 3,392 tweets were annotated twice by the same annotator, and 12,214 were annoted twice by two different annotators (Mozetič et al., 2016). These tweets as well as the ambiguous ones (with annotations by different hands) were discarded, so that 82,715 remained. Of these, only 59,107 were available at the time this paper was written, and just 10,639 (2,935 negative, 2,677 neutral, and 5,027 positive tweets) containing 624 different *emojis*.

[9] Available at `http://hdl.handle.net/11356/1054`.
[10] Available at `http://www.gamasystem.si`.
[11] Available at `http://www.perceptionanalytics.net`.

- The Spanish dataset, collected between May 2013 and December 2014, and composed of 275,588 tweets (33,978 negative, 107,467 neutral, and 134,143 positive tweets), manually labeled by 5 native speakers. Of these 40,205 tweets were annotated twice by the same annotator, and 2,194 were annotated by two different people (Mozetič et al., 2016), delivering the result of 233,189 unique tweets. Again, these tweets as well as the ambiguous ones (with annotations by different hands) were discarded, so that 214,948 remained. In this case, only 169,855 were available at the time this paper was written. The outcome was 12,759 tweets (1,022 negative, 3,431 neutral, and 8,306 positive tweets) containing 613 different *emojis*.

If we compare the manually labeled tweets with *emojis*, we obtain the distributions of negative, neutral, and positive tweets for the English and Spanish sets. Combining both datasets, there are 704 different *emojis* in them. These results are shown in Table 1.

| **Language** | **Different *emojis*** | **Sentiment** | **Tweets with *emojis*** | **Percentage** |
|---|---|---|---|---|
| English | 624 | Negative | 2,935 | 27.59% |
| | | Neutral | 2,677 | 25.16% |
| | | Positive | 5,027 | 47.25% |
| Spanish | 613 | Negative | 1,022 | 8.01% |
| | | Neutral | 3,431 | 26.89% |
| | | Positive | 8,306 | 65.10% |
| Spanish+English | 704 | Negative | 3,957 | 16.91% |
| | | Neutral | 6,108 | 26.11% |
| | | Positive | 13,333 | 56.98% |

Table 1: Distribution of negative, positive, and neutral tweets containing *emojis* for the English and Spanish sets.

### *4.2. Notation*

Before describing the results obtained for the Spanish and English datasets (and their union), we define the elements involved in the evaluation. These consist of two types of sets: the sets of tweets belonging to the datasets with sentiment values obtained after applying automatic USSPAD or tagging them manually, and the sets of *emojis* with their corresponding polarities[12]. In both cases, three sentiment values were considered: positive, neutral, and negative. Below we provide a more detailed explanation of these sets along with the evaluation results. Table 2 summarizes the notation of the sets involved in the evaluation.

The first set *sentiment values for texts* has two subsets:

- *annotated* sets represent the final sentiment results of each tweet taking into account the *emojis* included in a dataset manually annotated by (Kralj Novak et al., 2015), as shown in Figure 4a. The Spanish dataset is identified as $annotated_{es}$. The English dataset is identified as $annotated_{en}$ and their union is identified as $annotated_{es+en}$.

- *unsupervised* sets contain the final sentiment result of each tweet of a dataset after applying USSPAD (Fernández-Gavilanes et al., 2016), ignoring the *emojis*. This method uses the SO-CAL lexica together with AFINN. That is, unsupervised sets are automatically created, as shown in Figure 4b. The Spanish and English datasets are identified as $unsupervised_{es}$ and $unsupervised_{en}$, respectively. Finally, their union is called $unsupervised_{es+en}$. At this point, we should note that the unsupervised system does not require uniform distribution of the polarities because no training is performed.

Considering now the second set *sentiment values for emojis*, these sentiments are estimated in three ways:

[12]These are the *emoji sentiment lexica* developed, including both our approach from descriptions, its variants obtained from different combinations, and other approaches (Kralj Novak et al., 2015; Lu et al., 2016; Kimura & Katsurai, 2017).

| Set | Notation | Explanation |
|---|---|---|
| Set of sentiment values from texts | *annotated*$_{es}$ | Sentiment results obtained on the Spanish dataset after manual labelling by 5 annotators considering *emojis*. |
| | *annotated*$_{en}$ | Sentiment results obtained on the English dataset after manual labelling by 9 annotators considering *emojis*. |
| | *annotated*$_{es+en}$ | Sentiment results obtained on Spanish+English dataset after manual labelling by 14 annotators considering *emojis*. |
| | *unsupervised*$_{es}$ | Sentiment results obtained on Spanish dataset using USSPAD and ignoring *emojis*. |
| | *unsupervised*$_{en}$ | Sentiment results obtained on English dataset using USSPAD and ignoring *emojis*. |
| | *unsupervised*$_{es+en}$ | Sentiment results obtained on Spanish+English dataset using USSPAD ignoring *emojis*. |
| Set of sentiment values of *emojis* | *unsupervised*$_{emojiDef}$ | *Emoji* sentiment ranking estimated from *emoji* definitions through USSPAD. |
| | LIWC$_{emoji\text{CLDR}}$ | *Emoji* sentiment ranking estimated from *emoji* CLDR names using LIWC. |
| | *WordNet-Affect*$_{emoji}$ | *Emoji* sentiment ranking estimated from sentiment words using *WordNet-Affect*. |
| | $R_{annotated_{es}}$ | *Emoji* sentiment ranking estimated from *annotated*$_{es}$ set through formalization of sentiment. |
| | $R_{annotated_{en}}$ | *Emoji* sentiment ranking estimated from *annotated*$_{en}$ set through formalization of sentiment. |
| | $R_{annotated_{es+en}}$ | *Emoji* sentiment ranking estimated from *annotated*$_{es+en}$ set through formalization of sentiment. |
| | $R_{unsupervised_{es}}$ | *Emoji* sentiment ranking estimated from *unsupervised*$_{\text{es}}$ set through formalization of sentiment. |
| | $R_{unsupervised_{en}}$ | *Emoji* sentiment ranking estimated from *unsupervised*$_{\text{en}}$ set through formalization of sentiment. |
| | $R_{unsupervised_{es+en}}$ | *Emoji* sentiment ranking estimated from *unsupervised*$_{\text{es+en}}$ set through formalization of sentiment. |
| | $R_{annotated_{all}}$ | Publicly available *emoji* sentiment ranking. |

Table 2: Notation of the sets of sentiment values of texts and *emojis*.

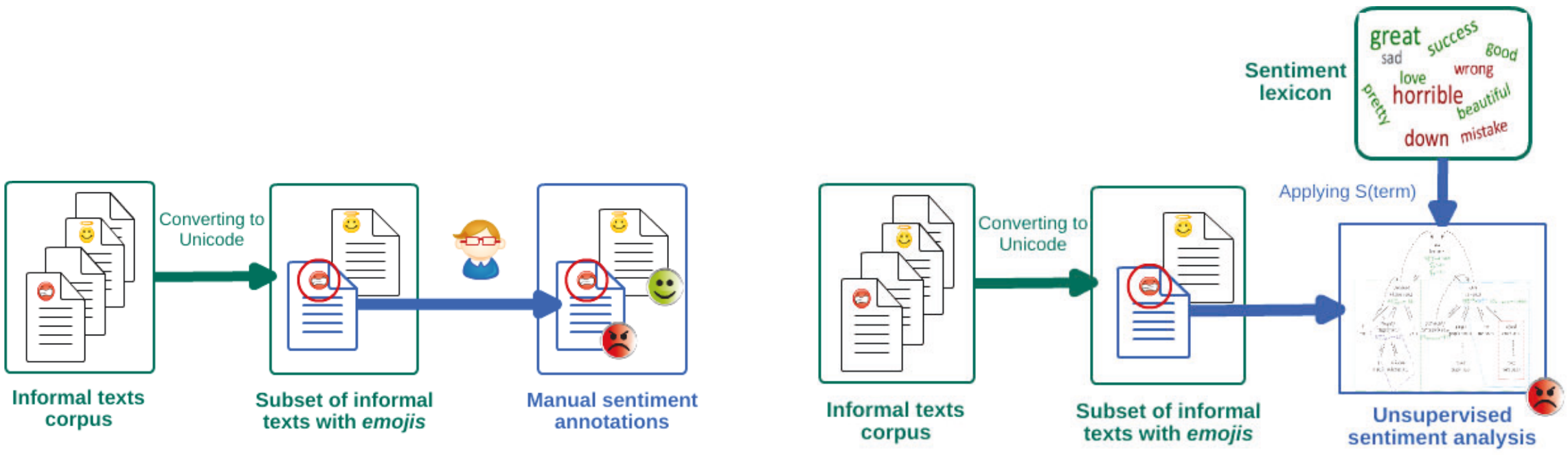

(a) Process to get *annotated* set.

(b) Process to get *unsupervised* set.

Figure 4: Processes to obtain *annotated* and *unsupervised* sets.

- The first method consists of applying our described approach, using USSPAD (with the same sentiment

lexica) on *emoji* definitions from `Emojipedia`, as shown in Figure 5a. The idea is to obtain an initial sentiment value for each *emoji* to be used later during SA of tweets. The result is a lexicon of *emojis* with polarities called $unsupervised_{emojiDef}$.

- The second method consists of applying the *emoji* lexicon obtained using the approach by (Lu et al., 2016), that is, applying LIWC to the short CLDR names of each *emoji*, as shown in Figure 5b. The result is a lexicon of *emojis* with polarities called $\text{LIWC}_{emoji\text{CLDR}}$.
- The third method consists of applying the *emoji* lexicon provided by the approach in (Kimura & Katsurai, 2017). It extracts sentiment words from *WordNet-Affect* and then calculates an *emoji* sentiment score vector based on co-occurrences between *emojis* and sentiment words. We call the resulting lexicon of *emojis* with polarities $WordNet\text{-}Affect_{emoji}$.
- The fourth method is described in (Kralj Novak et al., 2015). A ranking of a sentiment map of different *emojis* is estimated through a formalization. This formalization represents the sentiment of individual tweets, which can be negative, neutral, or positive.

  An *emoji*, whose assigned sentiment is represented by a 3-state variable $c \in \{-1, 0, +1\}$, may occur in different tweets. For this reason, for the set of relevant tweets, a discrete distribution may model the sentiment distribution by means of Eq.1:

$$\sum_{c} N(c) = N, c \in \{-1, 0, +1\}, \tag{1}$$

where $N$ denotes the number of occurrences of the *emojis*, while $N(c)$ are the occurrences of the *emojis* in tweets with sentiment labeled by $c$.

Thus, a discrete probability distribution can be obtained considering the three values. That is:

$$(p_{-1}, p_0, p_{+1}), \sum_{c} p_c = 1 \tag{2}$$

These probabilities are estimated from relative frequencies $p_c = \frac{N(c)}{N}$ when large samples are considered. When the samples are smaller, such as in our case, it is better to consider $p_c = \frac{N(c)+1}{N+k}$ (avoiding $p_c = 0$), where $k$ is the cardinality of the class (in our case $k = 3$).

Once the discrete probability distribution is defined, and following SentiWordNet (Baccianella et al., 2010), subjectivity is expressed as the opposite of neutrality. In other words, it can be expressed by the probabilities of negativity and positivity. Therefore, the sentiment score $\bar{s}$ is defined as the mean of the discrete probability distribution:

$$\bar{s} = -1 \cdot p_{-1} + 0 \cdot p_0 + 1 \cdot p_{+1} = -1 \cdot p_{-1} + 1 \cdot p_{+1}, \text{ where } -1 < \bar{s} < +1 \tag{3}$$

From this formalization, an *emoji* is assigned a sentiment from all the texts in which it occurs. Therefore, if these estimations are applied on the *emojis* coming from $annotated_{es}$, $annotated_{en}$, or $annotated_{es+en}$, the resulting lexica of *emoji* sentiments are $R_{annotated_{es}}$, $R_{annotated_{en}}$, and $R_{annotated_{es+en}}$, respectively. Figure 5c illustrates this process. Similarly, if the estimations are applied on $unsupervised_{es}$, $unsupervised_{en}$, or $unsupervised_{es+en}$, as shown in Figure 5d, the lexica of *emoji* sentiments are called $R_{unsupervised_{es}}$, $R_{unsupervised_{en}}$, and $R_{unsupervised_{es+en}}$, respectively.

In study by (Kralj Novak et al., 2015), the resulting lexicon of *emoji* sentiments, termed $R_{annotated_{all}}$, is that estimated from 13 language datasets[13]. That is, user language/country is not taken into account.

[13] Available at `https://www.clarin.si/repository/xmlui/bitstream/handle/11356/1048/Emoji_Sentiment_Data_v1.0.csv?sequence=8&isAllowed=y`.

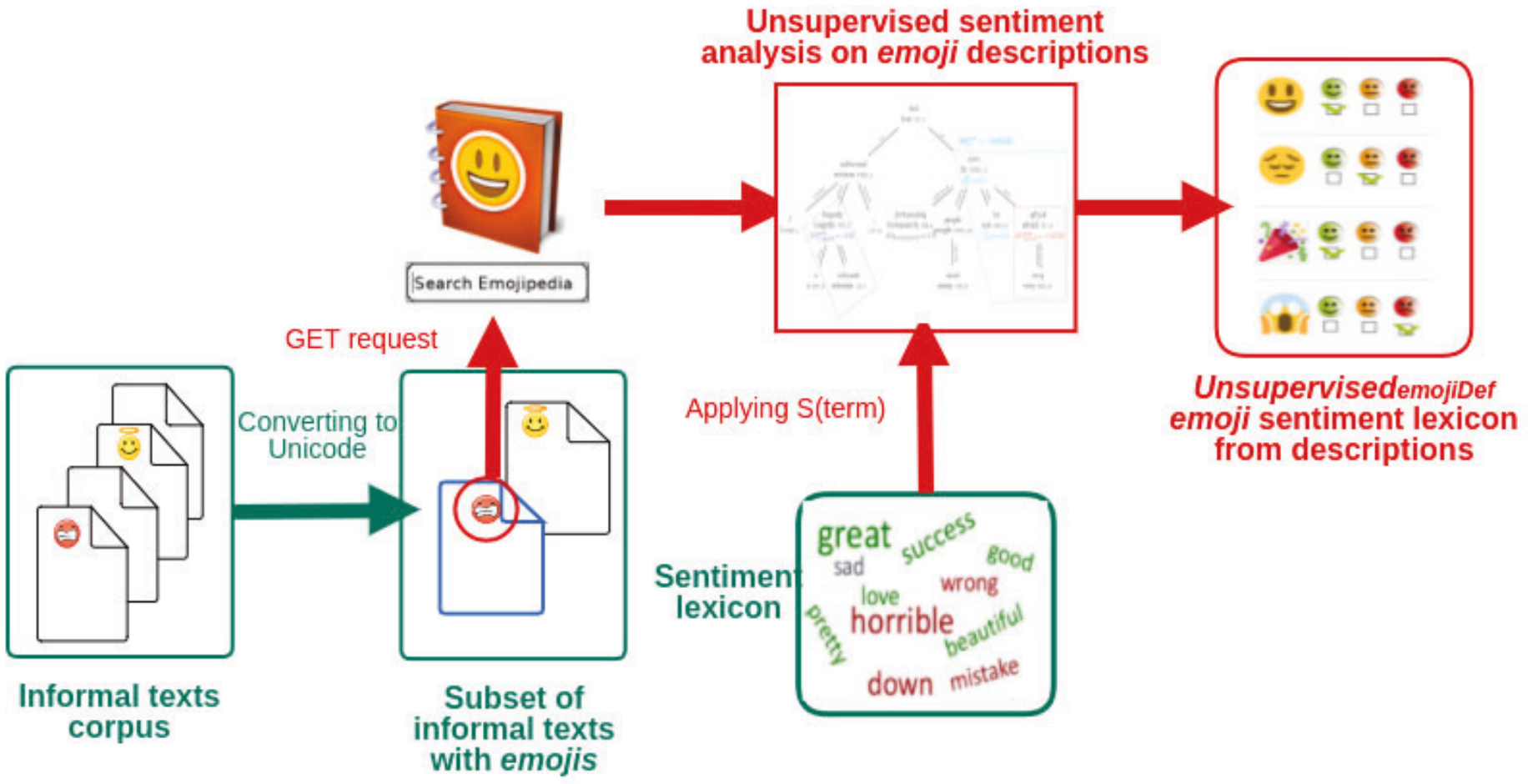

(a) Procedure for obtaining the $unsupervised_{emojiDef}$ lexicon.

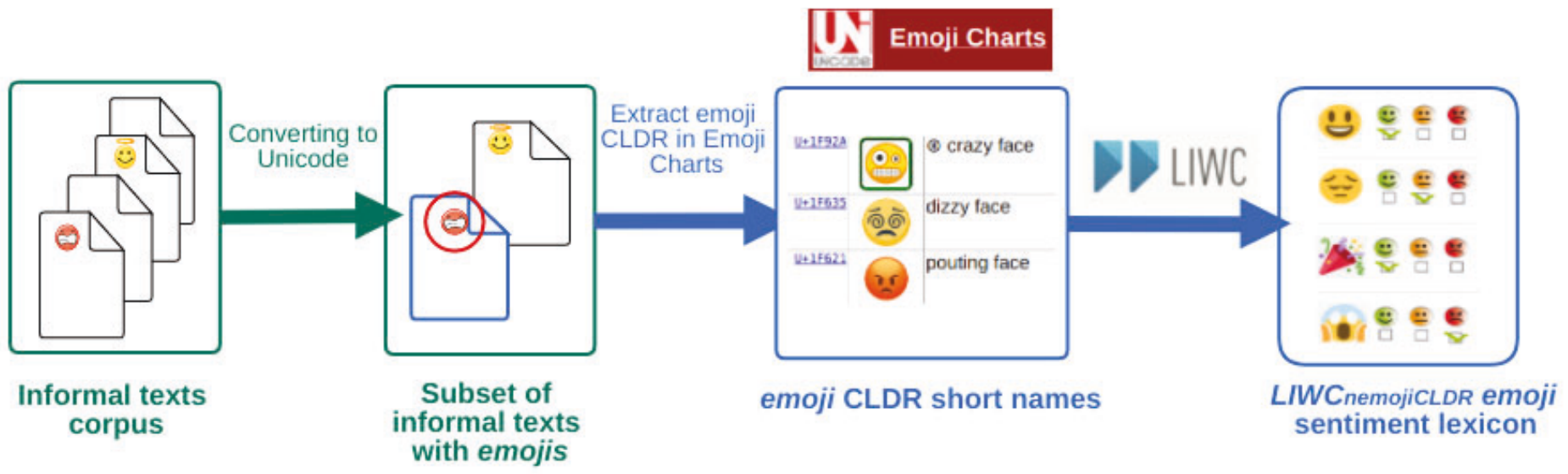

(b) Procedure for obtaining the LIWC$_{emoji\text{CLDR}}$ lexicon.

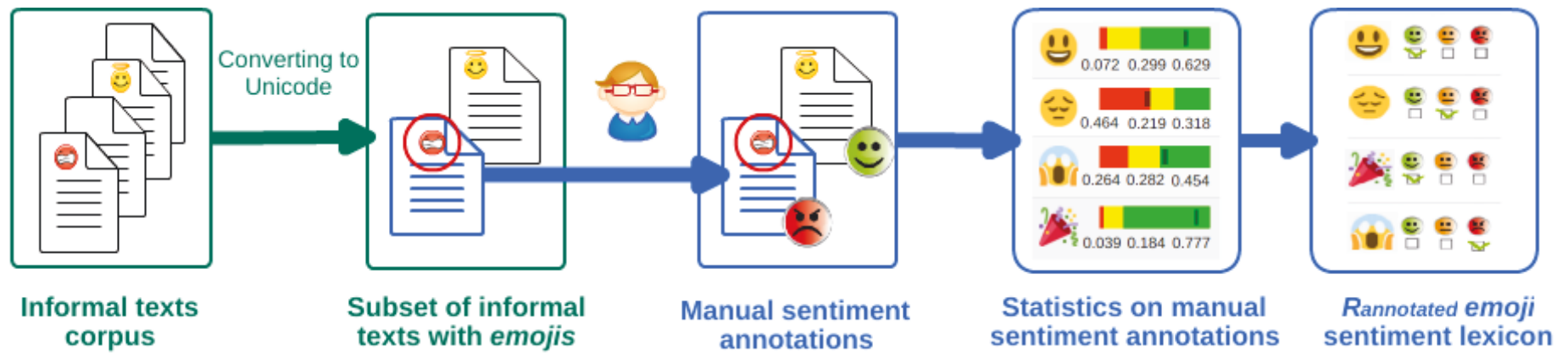

(c) Procedure for obtaining the $R_{annotated}$ lexicon.

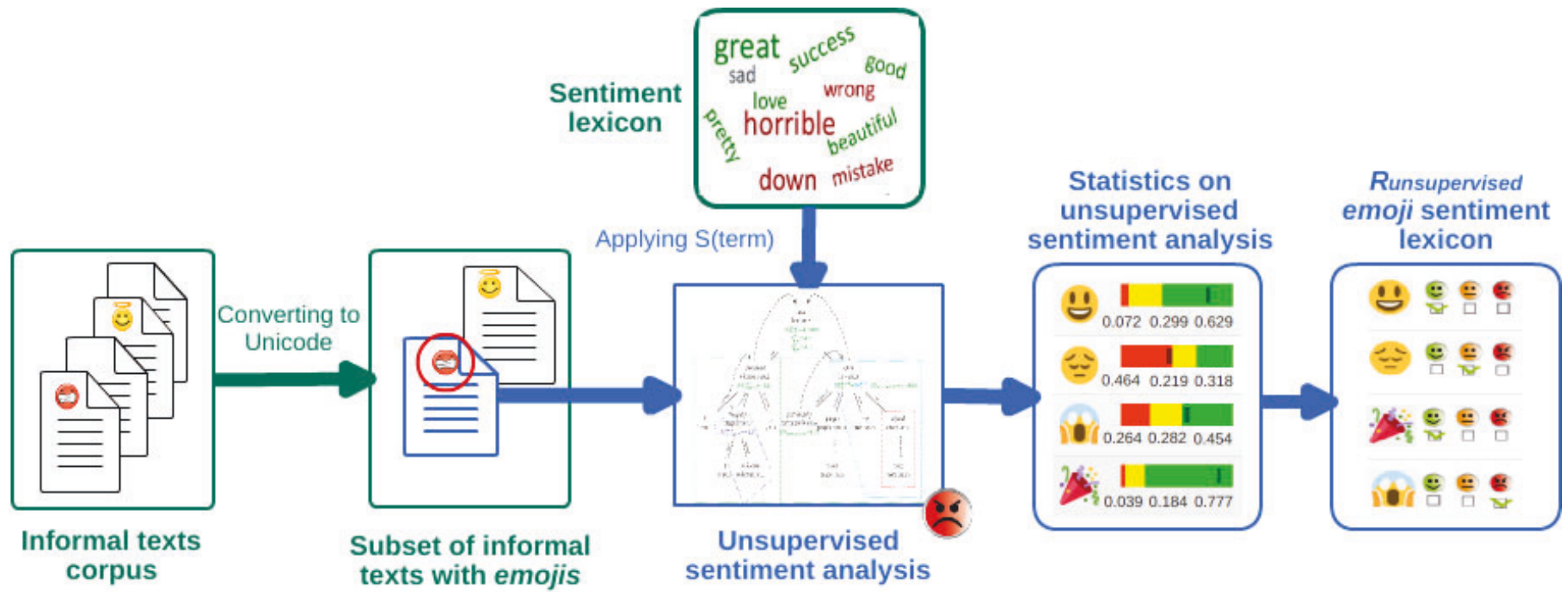

(d) Procedure for obtaining the $R_{unsupervised}$ lexicon.

Figure 5: Procedure for obtaining the lexica of *emoji* sentiments.

### 4.3. Results of our approach

To evaluate the impact of *emoji* descriptions, we take the datasets described in (Kralj Novak et al., 2015) and (Mozetič et al., 2016) as gold standard, because they have been manually labelled. We employ accuracy as a straightforward SA performance metric to check the level of agreement of our outcome with test data. However, due to the low robustness of this metric in case of class imbalance (Sokolova & Lapalme, 2009; Rossi et al., 2016), we also applied more sofisticated SA metrics: the precision ($P_{macro}$) and $F$ ($F_{macro}$) macroaverages, calculating them on the positive and negative classes. Then, we applied them to three different tweet datasets (Spanish, English and mixed) using USSPAD, which has amply demonstrated its performance (Fernández-Gavilanes et al., 2016). The aim of this evaluation is not to produce state-of-the-art SA results, but simply to show that including *emojis* and their descriptions adds discriminating information that could be exploited in more advanced social NLP systems, such as in USSPAD. For this reason, it is necessary to determine if these definitions provide relevant information (in comparison with, for example, CLDR short names) when manual annotation is not feasible.

| **Approach** | | | **Acc** | $P_{macro}$ | $F_{macro}$ |
|---|---|---|---|---|---|
| **Without *emojis*** | (*A1*) | $unsupervised_{es}$ | 53.13% | 43.48% | 42.09% |
| | | $unsupervised_{en}$ | 53.15% | 50.42% | 57.08% |
| | | $unsupervised_{es+en}$ | 53.14% | 51.91% | 54.50% |
| **Lu et al., 2016** | (*A2*) | $unsupervised_{es}$ + $\text{LIWC}_{emoji\text{CLDR}}$ | 55.12% | 48.46% | 44.49% |
| | | $unsupervised_{en}$ + $\text{LIWC}_{emoji\text{CLDR}}$ | 57.82% | 62.27% | 63.97% |
| | | $unsupervised_{es+en}$ + $\text{LIWC}_{emoji\text{CLDR}}$ | 56.35% | 60.14% | 58.62% |
| **Kimura et al., 2017** | (*A3*) | $unsupervised_{es}$ + $WordNet\text{-}Affect_{emoji}$ | 61.41% | 50.65% | 46.62% |
| | | $unsupervised_{en}$ + $WordNet\text{-}Affect_{emoji}$ | 55.63% | 61.67% | 63.45% |
| | | $unsupervised_{es+en}$ + $WordNet\text{-}Affect_{emoji}$ | 58.78% | 58.57% | 59.32% |
| **Our approach** | (*A4*) | $unsupervised_{es}$ + $unsupervised_{emojiDef}$ | 56.49% | 50.00% | 44.93% |
| | | $unsupervised_{en}$ + $unsupervised_{emojiDef}$ | 59.82% | 67.51% | 66.57% |
| | | $unsupervised_{es+en}$ + $unsupervised_{emojiDef}$ | 58.00% | 63.17% | 60.11% |
| **Novak et al., 2015** | (*A5*) | $unsupervised_{es}$ + $R_{annotated_{all}}$ | 60.36% | 51.51% | 46.32% |
| | | $unsupervised_{en}$ + $R_{annotated_{all}}$ | 60.43% | 70.06% | 68.74% |
| | | $unsupervised_{es+en}$ + $R_{annotated_{all}}$ | 60.39% | 64.92% | 62.86% |

Table 3: Accuracy and macroaveraging metrics using USSPAD.

Table 3 shows the results for the Spanish, English, and Spanish+English datasets. We applied USSPAD, our unsupervised system with sentiment propagation across dependencies, in four different ways:

- ($A1$) ignoring the polarities of *emojis*, as shown in Figure 4b;
- ($A2$) considering the *emoji* lexicon obtained by applying LIWC on *emoji* CLDR names, as illustrated in Figure 6;
- ($A3$) considering the *emoji* lexicon provided by (Kimura & Katsurai, 2017), obtained by applying *WordNet-Affect* on tweets, as illustrated in Figure 7;
- ($A4$) considering the *emoji* lexicon resulting from their descriptions (our approach), as illustrated in Figure 2;
- ($A5$) including the publicly available values for *emoji* sentiments estimated from the labels of the 13 language datasets in (Kralj Novak et al., 2015), as shown in Figure 8.

Table 3 shows the results of the four approaches[14] applied in combination with *unsupervised* datasets. Exploiting *emoji* CLDR names, as described in (Lu et al., 2016), *WordNet-Affect* entries, as described in (Kimura & Katsurai, 2017) and *emoji* definitions improves USSPAD accuracy between 2%-4%, 2%-8% and 4%-7%, respectively. Considering definitions (instead of CLDR short names) and semantic dependencies such as negations (instead of isolated words with LIWC) also improves the results.

[14]We tried to contact the authors of the approach described in (Hussien et al., 2016) to obtain the *emoji* lexicon (because it could not be replicated from that study), but there was no reply.

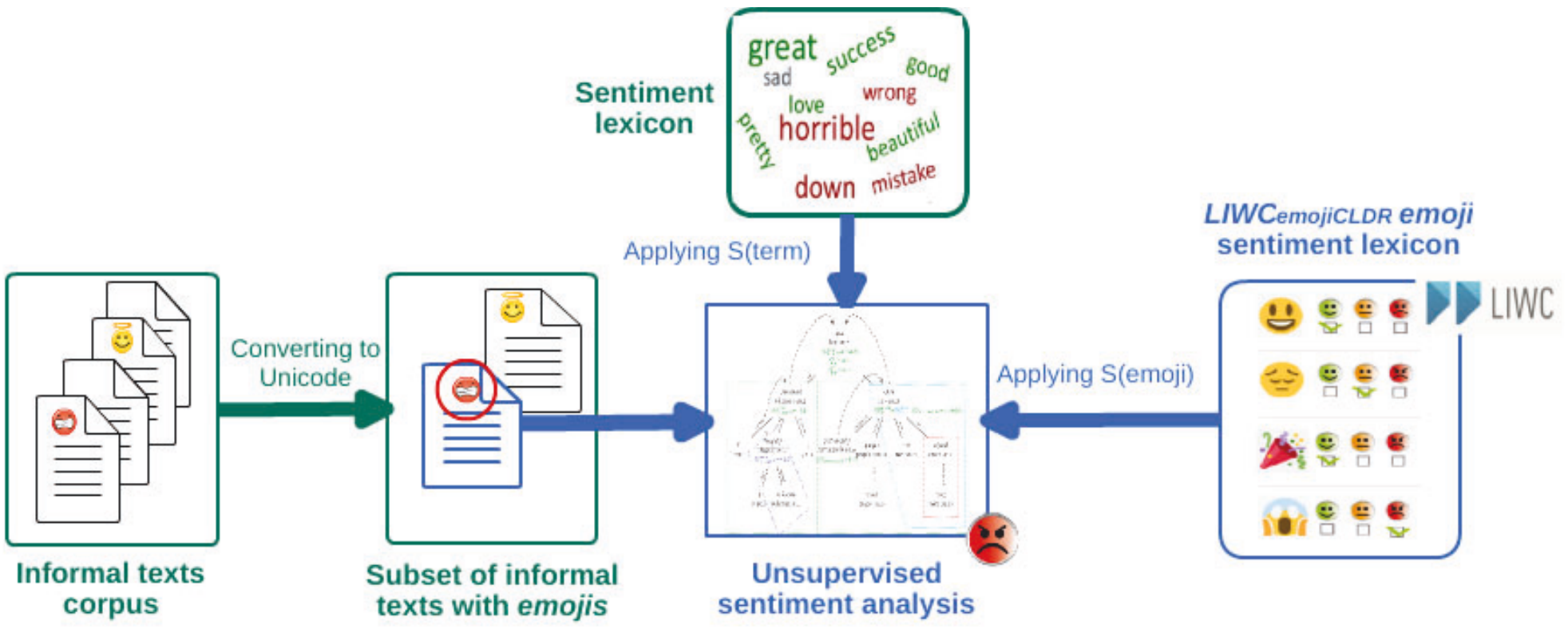

Figure 6: Approach using the LIWC$_{emoji\text{CLDR}}$ lexicon with the *unsupervised* set.

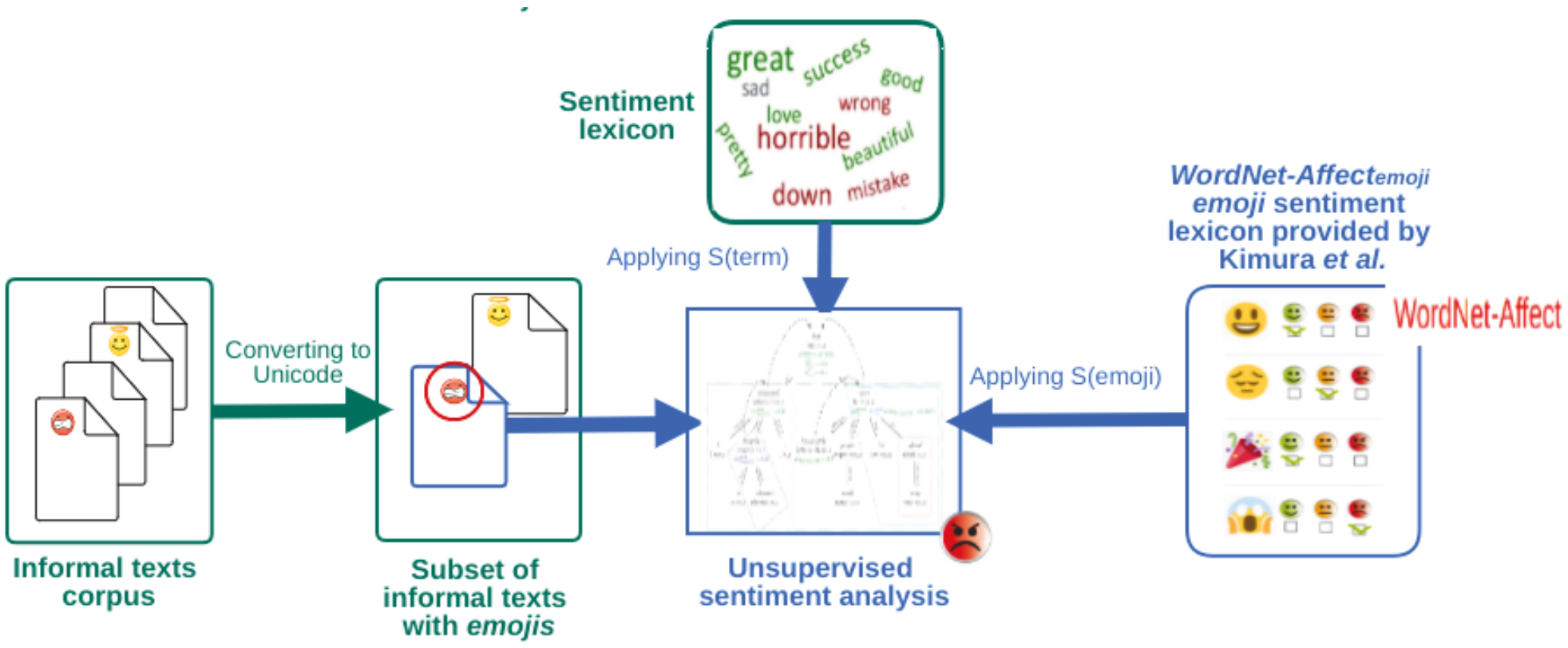

Figure 7: Approach using the *WordNet-Affect*$_{emoji}$ lexicon with the *unsupervised* set.

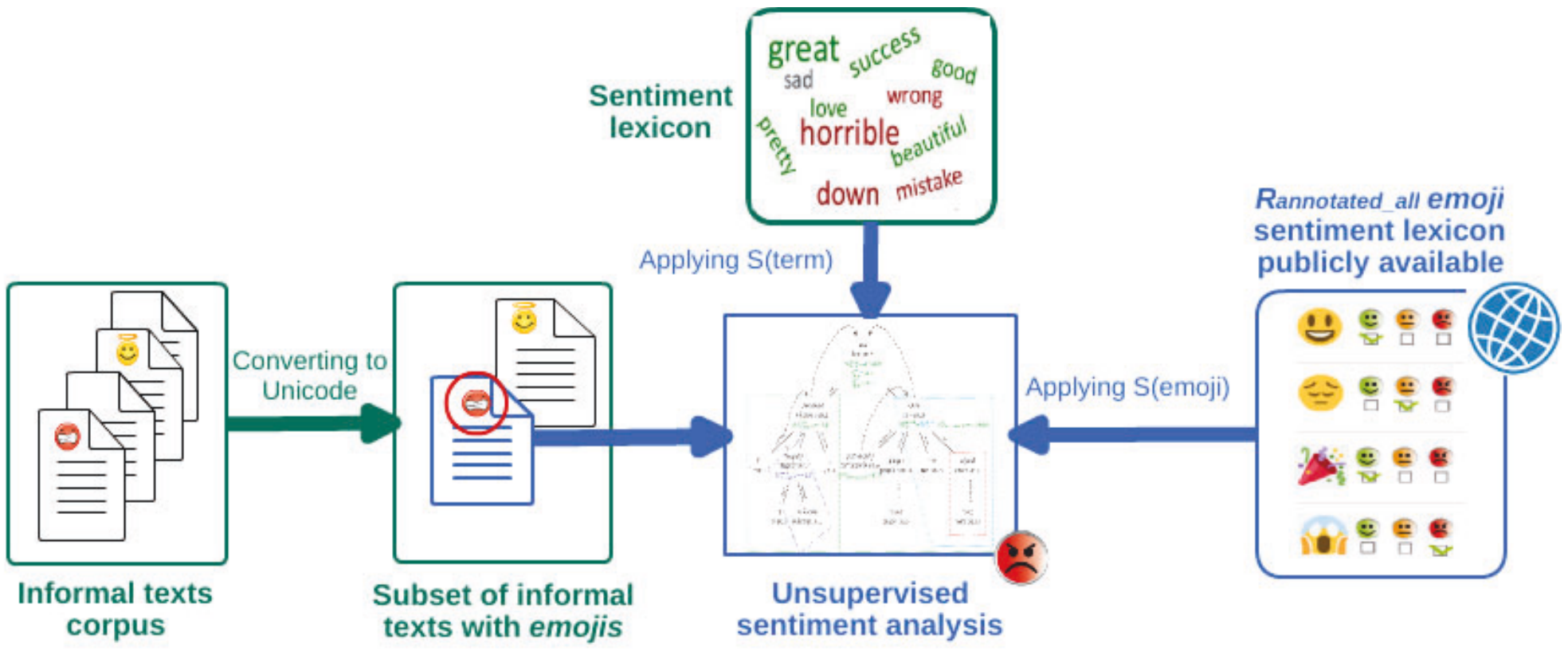

Figure 8: Approach using the $R_{annotated_{all}}$ lexicon with the *unsupervised* set.

If the comparison metrics are $P_{macro}$ and $F_{macro}$, the improvements achieved depending on the strategy and the language are significant. For example, the improvements over USSPAD by exploiting *emoji* CLDR short names or *WordNet-Affect* entries are 5%-11% for $P_{macro}$ and 2%-6% for $F_{macro}$. However, when *emoji* definitions are applied in USSPAD, these improvements are 7%-17% for $P_{macro}$ and 3%-10% for $F_{macro}$.

At the same time, these results show that *emoji* values provided by external agents, such as the ranking by (Kralj Novak et al., 2015) ($R_{annotated_{all}}$) only result in an additional extra improvement of between 1% and 3% (for all the metrics considered, accuracy, $P_{macro}$ and $F_{macro}$), depending on the language, with respect to our approach.

Unlike the case of the $R_{annotated_{all}}$ (Kralj Novak et al., 2015) lexicon, with time-consuming manual labelling of tweets, our approach delivers results in a speedy unsupervised way and does not depend on external resources such as *WordNet*. The results clearly indicate that the less time consuming USSPAD combined with *emoji* definitions is a good starting point to achieve an efficient *emoji* sentiment lexicon.

### *4.4. Results of our enhanced approach*

Now we are in a position to apply three different unsupervised approaches, two of them variants of the approach described in the previous section. In all cases, extra information is incorporated during the propagation, i.e. *emoji* sentiment values obtained in different ways. Each evolution adds, to the *unsupervised* sets, extra information values obtained from the estimations described in (Kralj Novak et al., 2015), which respectively result from:

($E1$) the corresponding ranking of the sentiment map applied to the *unsupervised* sets (and not to the labelled data), $R_{unsupervised}$,

($E2$) the estimations, denotated as $R_{(unsupervised+unsupervised_{emojiDef})}$, calculated after applying our approach to informal texts,

($E3$) the mean of the values obtained from $unsupervised_{emojiDef}$ and from the lexicon used in the second variant on each *emoji*, represented as $[\overline{unsup._{emojiDef} + R_{(unsup. + unsup._{emojiDef})}}]$.

For the sake of clarity, Figure 9a shows the way $R_{unsupervised}$ sets are obtained using USSPAD . Figure 9b shows statistics obtained after applying USSPAD to tweets to get $R_{(unsupervised+unsupervised_{emojiDef})}$. Finally, Figure 9c shows how, once lexica $unsupervised_{emojiDef}$ and $R_{(unsupervised+unsupervised_{emojiDef})}$ are obtained, the results are averaged for each *emoji* in order to get the $[\overline{unsup._{emojiDef} + R_{(unsup.+unsup._{emojiDef})}}]$ lexicon. Moreover, in order to compare our enhanced unsupervised approaches, we include two variants of the work in (Kralj Novak et al., 2015), with manual labelling. Respectively,

($E4$) the publicly available values of *emoji* sentiments estimated from the labels of the 13 language datasets, and

($E5$) the corresponding estimation from the *annotated* set.

In both cases, *emoji* sentiment is estimated from the manual labels of all tweets in which a particular *emoji* is included. This process takes a considerable amount of time.

Table 4, 5 and 6 show the results for the English, Spanish, and Spanish+English datasets, respectively.

#### *4.4.1. Results for the English dataset*

| **Approach** | | **Acc** | $P_{macro}$ | $F_{macro}$ |
|---|---|---|---|---|
| **Our approach** | $unsupervised_{en}+unsupervised_{emojiDef}$ | 59.82% | 67.51% | 66.57% |
| | **(*E*1)** $unsupervised_{en}+R_{unsupervised_{en}}$ | 60.46% | 68.65% | 67.29% |
| | **(*E*2)** $unsupervised_{en}+R_{(unsupervised_{en}+unsupervised_{emojiDef})}$ | 61.57% | 74.28% | 70.18% |
| | **(*E*3)** $unsupervised_{en}+[\overline{unsup._{emojiDef} + R_{(unsup._{en}+unsup._{emojiDef})}}]$ | 61.70% | 73.28% | 69.62% |
| **Novak et al., 2015** | **(*E*4)** $unsupervised_{en}+R_{annotated_{all}}$ | 60.43% | 70.06% | 68.74% |
| | **(*E*5)** $unsupervised_{en}+R_{annotated_{en}}$ | 63.76% | 76.52% | 72.17% |

Table 4: Accuracy and macroveraging metrics. Unsupervised sentiment analysis on the English dataset.

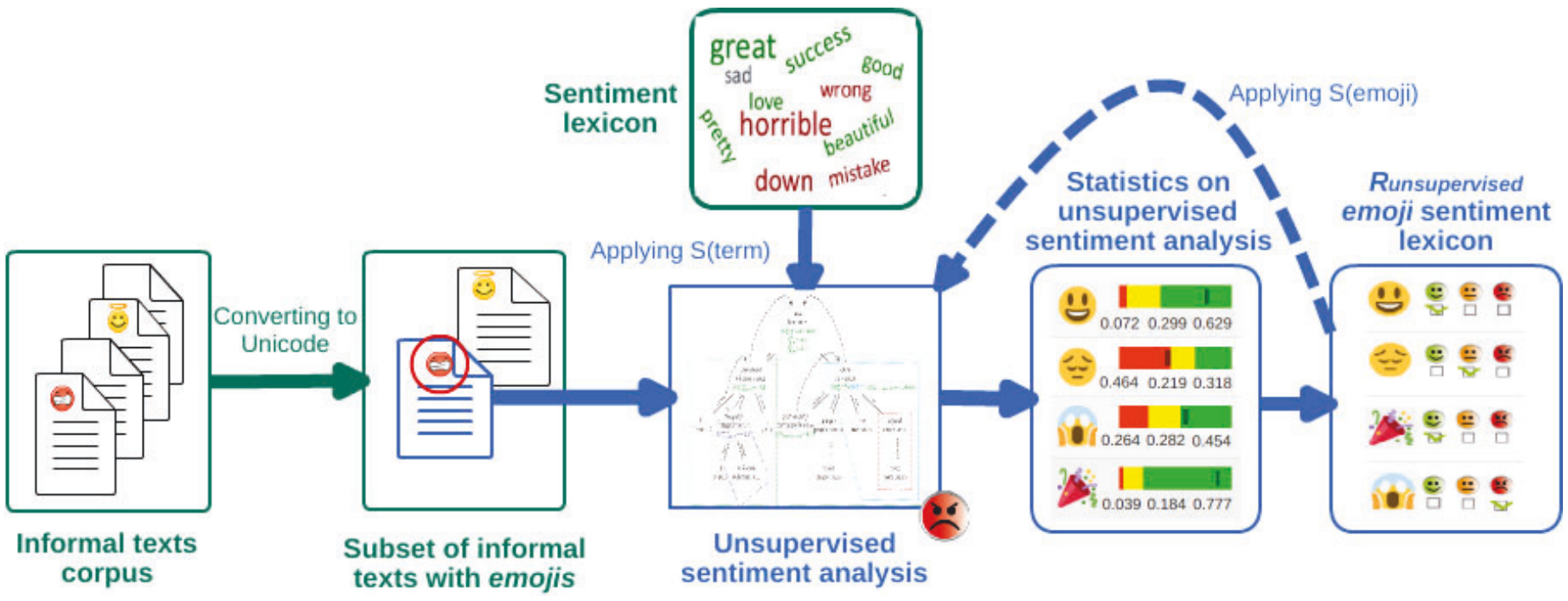

(a) Results using the $R_{unsupervised}$ lexicon.

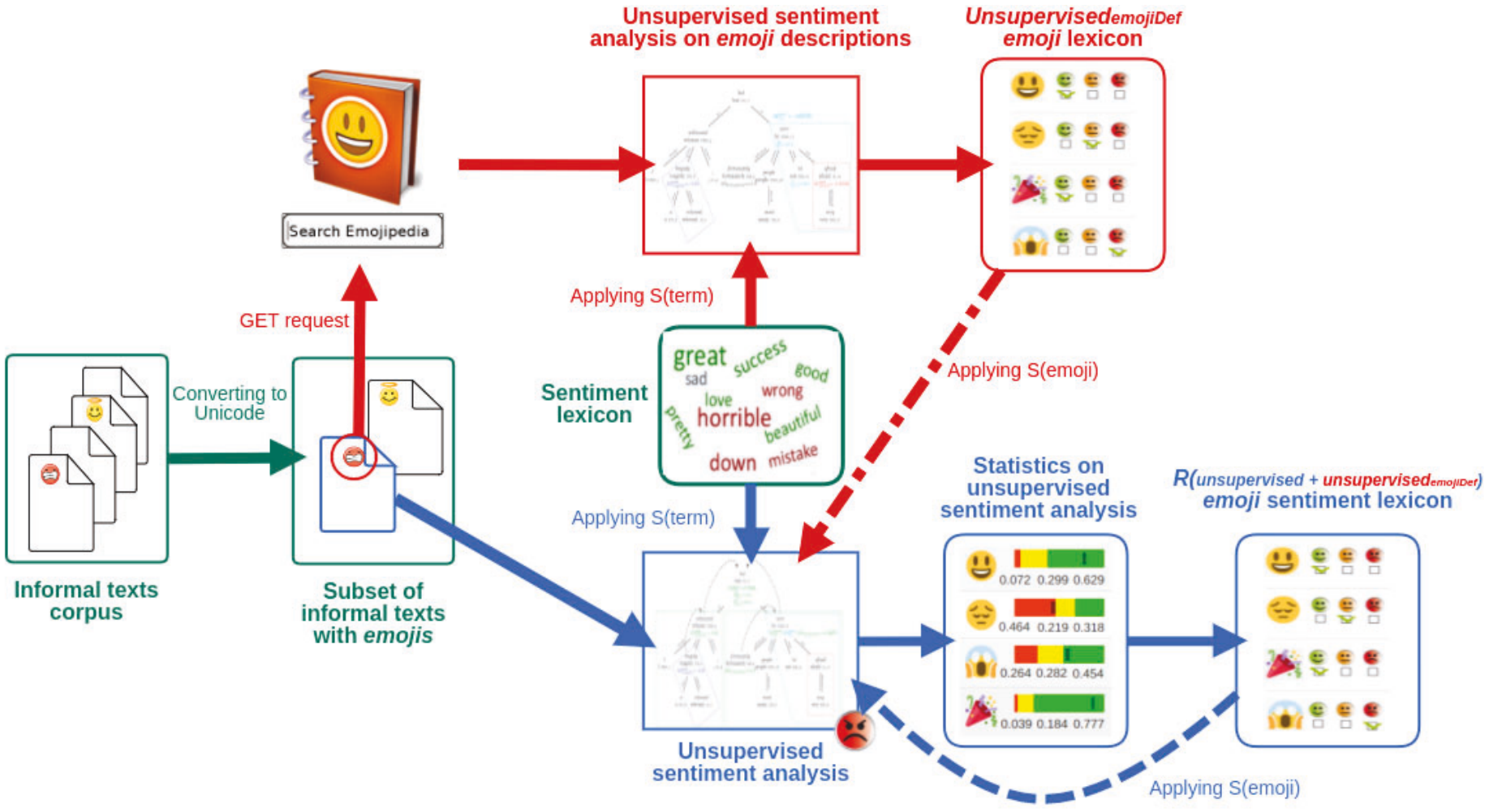

(b) Results using the $R_{(unsupervised+unsupervised_{emojiDef})}$ lexicon.

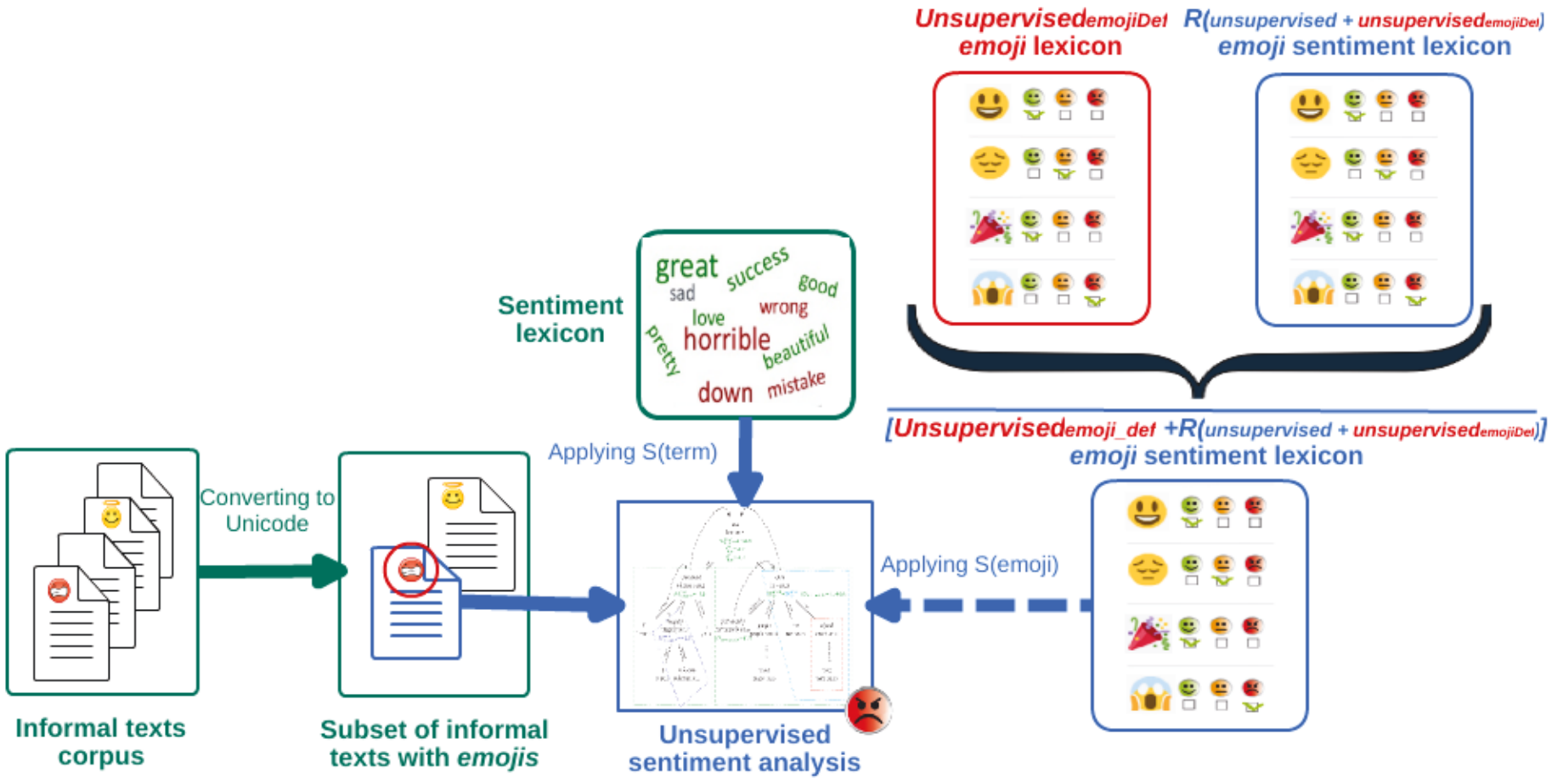

(c) Results using the $\overline{[unsup._{emojiDef} + R_{(unsup.+unsup._{emojiDef})}]}$ lexicon.

Figure 9: Results using variants of the *emoji* sentiment lexicon.

Table 4 shows the results of the approaches applied in combination with $unsupervised_{en}$. The first three variants outperform $unsupervised_{en}$ without *emojis* by 8%-9% in accuracy, 18%-23% in $P_{macro}$ and 10%-13% in $F_{macro}$. If we compare these same variants with $unsupervised_{en}$ + LIWC$_{emojiCLDR}$ of (Lu et al., 2016) in Table 3, the improvements are 3%-4% in accuracy, 5%-12% in $P_{macro}$ and 3%-6% in $F_{macro}$. Finally, if we compare them with $unsupervised_{en}$ + *WordNet-Affect*$_{emoji}$ (Kimura & Katsurai, 2017), the improvements are 5%-6% in accuracy, 7%-13% in$P_{macro}$ and 4%-7% in $F_{macro}$.

This shows that obtaining the sentiment values using the ranking from our approach in informal texts (introducing *emoji* definitions into the USSPAD) (E2) performed better for the English dataset than simply applying formalization on the sentiment resulting from $unsupervised_{en}$ (E1). Again, this clearly indicates that individual values coming from USSPAD combined with *emoji* definitions result in a good *emoji* sentiment lexicon. Moreover, if both the $unsupervised_{emojiDef}$ and $R_{(unsupervised_{en}+unsupervised_{emojiDef})}$ lexica are applied in equal conditions (E3), accuracy, $P_{macro}$ and $F_{macro}$ improve even futher.

Regarding the last two variants based on *emojis* labels provided by external agents, such as the ranking by (Kralj Novak et al., 2015) ($R_{annotated_{all}}$ (E4)) and estimations using formalization on the $annotated_{en}$ labels (E5), the improvement with respect to $unsupervised_{en}$ set without *emojis* is 7%-10% in accuracy, 19%-25% in $P_{macro}$ and 11%-5% in $F_{macro}$. It is worth nothing that the second and third variants achieve better performances than the fourth variant. Clearly, language is a key factor in the interpretation of the meaning of an *emoji*. Another interesting aspect is that the difference between the values obtained with the first three variants is much less than could be expected in comparison with the results of the last two variants, bearing in mind that the latter is based directly on the manual labels estimated for the English dataset ($R_{annotated_{en}}$).

#### *4.4.2. Results for the Spanish dataset*

| **Approach** | | **Accuracy** | $P_{macro}$ | $F_{macro}$ |
|---|---|---|---|---|
| **Our approach** | $unsupervised_{es}+unsupervised_{emojiDef}$ | 56.49% | 50.00% | 44.93% |
| | (***E1***) $unsupervised_{es}+R_{unsupervised_{es}}$ | 62.79% | 50.29% | 46.27% |
| | (***E2***) $unsupervised_{es}+R_{(unsupervised_{es}+unsupervised_{emojiDef})}$ | 62.52% | 50.72% | 46.56% |
| | (***E3***) $unsupervised_{es}+[unsup._{emojiDef} + R_{(unsup._{es}+unsup._{emojiDef})}]$ | 60.89% | 51.18% | 46.00% |
| **Novak et al., 2015** | (***E4***) $unsupervised_{es}+R_{annotated_{all}}$ | 60.36% | 51.51% | 46.32% |
| | (***E5***) $unsupervised_{es}+R_{annotated_{es}}$ | 63.12% | 50.43% | 46.86% |

Table 5: Accuracy and macroveraging metrics. Unsupervised sentiment analysis on the Spanish dataset.

Table 5 shows the results of the variants applied in combination with $unsupervised_{es}$. As could be expected, the first three variants outperform $unsupervised_{es}$ without *emojis* by 8%-10% in accuracy, 6%-7% in $P_{macro}$ and 4%-5% in $F_{macro}$. If we compare them with $unsupervised_{es}$ + LIWC$_{emojiCLDR}$, the improvements are 6%-8% in accuracy and 2%-3% in $P_{macro}$ and $F_{macro}$, although the improvements are lower than those for the English dataset. However, none of the three variants outperformed $unsupervised_{es}$ + *WordNet-Affect*$_{emoji}$ for all metrics at the same time. In terms of accuracy and $P_{macro}$, our second and third variants are better by about 1%-2%, but they are worse in terms of $F_{macro}$ (although the values are very close).

Moreover, the ranking of the first three variants changes in comparison with the English dataset. Against all odds, the first variant occupies the third place, whereas the third variant is ranked first. Moreover, there is a significant difference with the results obtained for $P_{macro}$ and $F_{macro}$ in this dataset in comparison with the English case. This may be caused by the poor quality of the annotation of the Spanish dataset. Most of this dataset (over 95%) was annotated by only one annotator. In (Mozetič et al., 2016), reliable estimations of this effect are provided (low values of self-agreement and inter-agreement).

Unlike what happens with the English dataset, for the two variants based on the *emoji* sentiments in (Kralj Novak et al., 2015) ($R_{annotated_{all}}$ (E4) and $R_{annotated_{es}}$ (E5)), the improvements with respect to $unsupervised_{es}$ without *emojis* are 7%-10% in accuracy, 6%-8% in $P_{macro}$ and 4% in $F_{macro}$. The first and second variants achieve better accuracies (respectively, the second and third variants better $P_{macro}$ and the second one better $F_{macro}$) than the fourth variant. As for the previous dataset, creating an *emoji* sentiment lexicon from texts in a given language seems highly advantageous.

#### 4.4.3. Results for the Spanish+English dataset

| Approach | | Acc | $P_{macro}$ | $F_{macro}$ |
|---|---|---|---|---|
| **Our approach** | $unsupervised_{es+en}$+$unsupervised_{emojiDef}$ | 58.00% | 63.17% | 60.11% |
| | (***E*1**) $unsupervised_{es+en}$+$R_{unsupervised_{es+en}}$ | 59.28% | 59.93% | 60.09% |
| | (***E*2**) $unsupervised_{es+en}$+$R_{(unsupervised_{es+en}+unsupervised_{emojiDef})}$ | 59.77% | 64.48% | 61.74% |
| | (***E*3**) $unsupervised_{es+en}$+$\boxed{unsup._{emojiDef} + R_{(unsup._{es+en}+unsup._{emojiDef})}}$ | 60.60% | 67.45% | 63.58% |
| **Novak et al., 2015** | (***E*4**) $unsupervised_{es+en}$+$R_{annotated_{all}}$ | 60.39% | 64.92% | 62.86% |
| | (***E*5**) $unsupervised_{es+en}$+$R_{annotated_{es+en}}$ | 60.39% | 63.77% | 62.78% |

Table 6: Accuracy and macroaveraging metrics. Unsupervised sentiment analysis on the Spanish+English dataset.

Table 6 shows the results of our variants applied in combination with $unsupervised_{es+en}$. The first three variants still outperform $unsupervised_{es+en}$ without *emojis*. The improvements are 5%-8% in accuracy, 8%-15% in $P_{macro}$ and 6%-10% in $F_{macro}$. The same occurs when the first three variants are compared with $unsupervised_{es+en}$ + LIWC$_{emoji\text{CLDR}}$, yielding improvements of 2%-5% in accuracy and $F_{macro}$. However, there is only improvement in $P_{macro}$ for the last two variants (5%-7%). If we compare our variants with $unsupervised_{es+en}$ + *WordNet-Affect*$_{emoji}$, the first three are better. The improvement in accuracy is low, of 1%-2%, but for $P_{macro}$ and $F_{macro}$ it reaches 2%-8% and 1%-5%, respectively.

Unlike what happens with the Spanish and English datasets, the last two variants, based on formalization of $R_{annotated_{es+en}}$ labels, perform worse than the independent datasets. Again, this may be due to the low quality of the Spanish dataset (Mozetič et al., 2016). However, if we compare the third variant of our approach with results obtained in (E4) and (E5), our approach achieves better results.

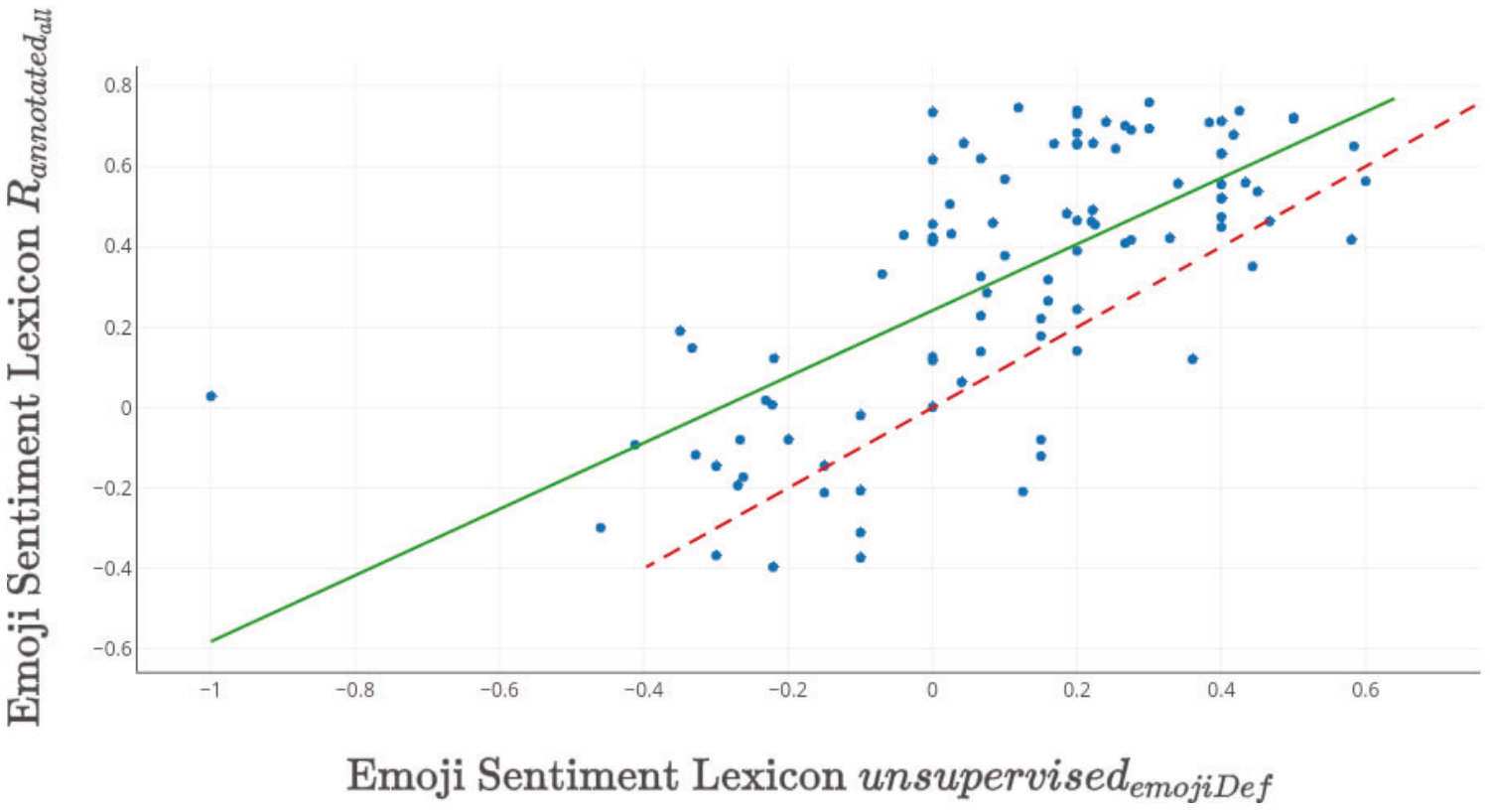

Figure 10: Scatterplot for top 100 *emoji* sentiment scores comparing $unsupervised_{emojiDef}$ with $R_{annotated_{all}}$

#### 4.5. Correlations between emoji sentiment lexica

In addition to evaluating the impact of *emoji* descriptions on the advanced USSPAD NLP system, we investigated the correlation between the conventional $R_{annotated_{all}}$ lexicon and our *emoji* sentiment lexica for positive, negative and neutral labels. For the $R_{annotated_{all}}$ lexicon a limited *emoji* subset was evaluated using a small number of tweets. Therefore, for a fair analysis, we only used the top 100 occurring *emojis* in $annotated_{all}$ to measure the correlation. In the following figures, our approaches are represented with dots, whereas the associated regression trendlines are shown as solid green lines. Red dashed lines represent the case for $R_{annotated_{all}}$.

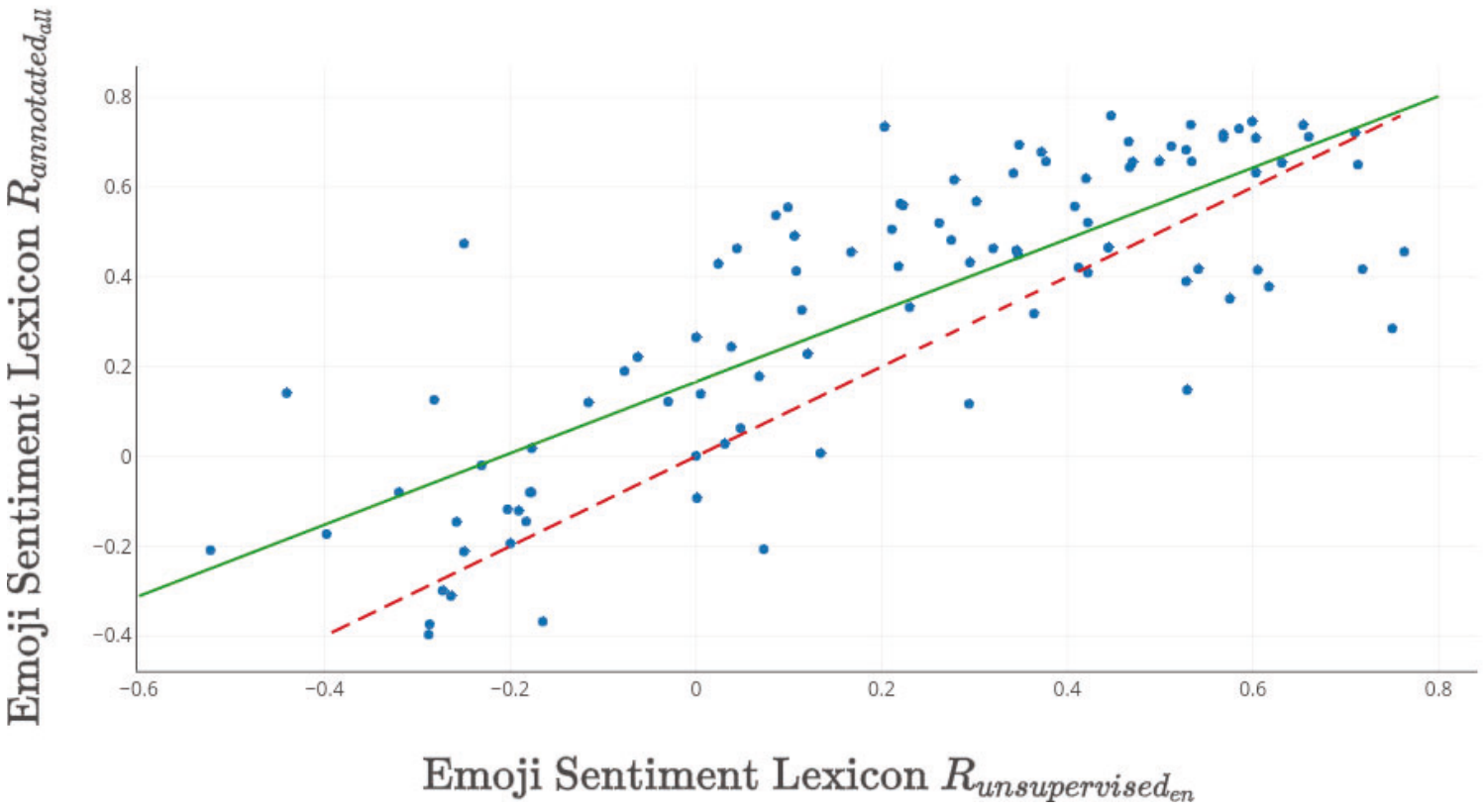

Figure 11: Scatterplot for top 100 *emoji* sentiment scores comparing $R_{unsupervised_{en}}$ with $R_{annotated_{all}}$

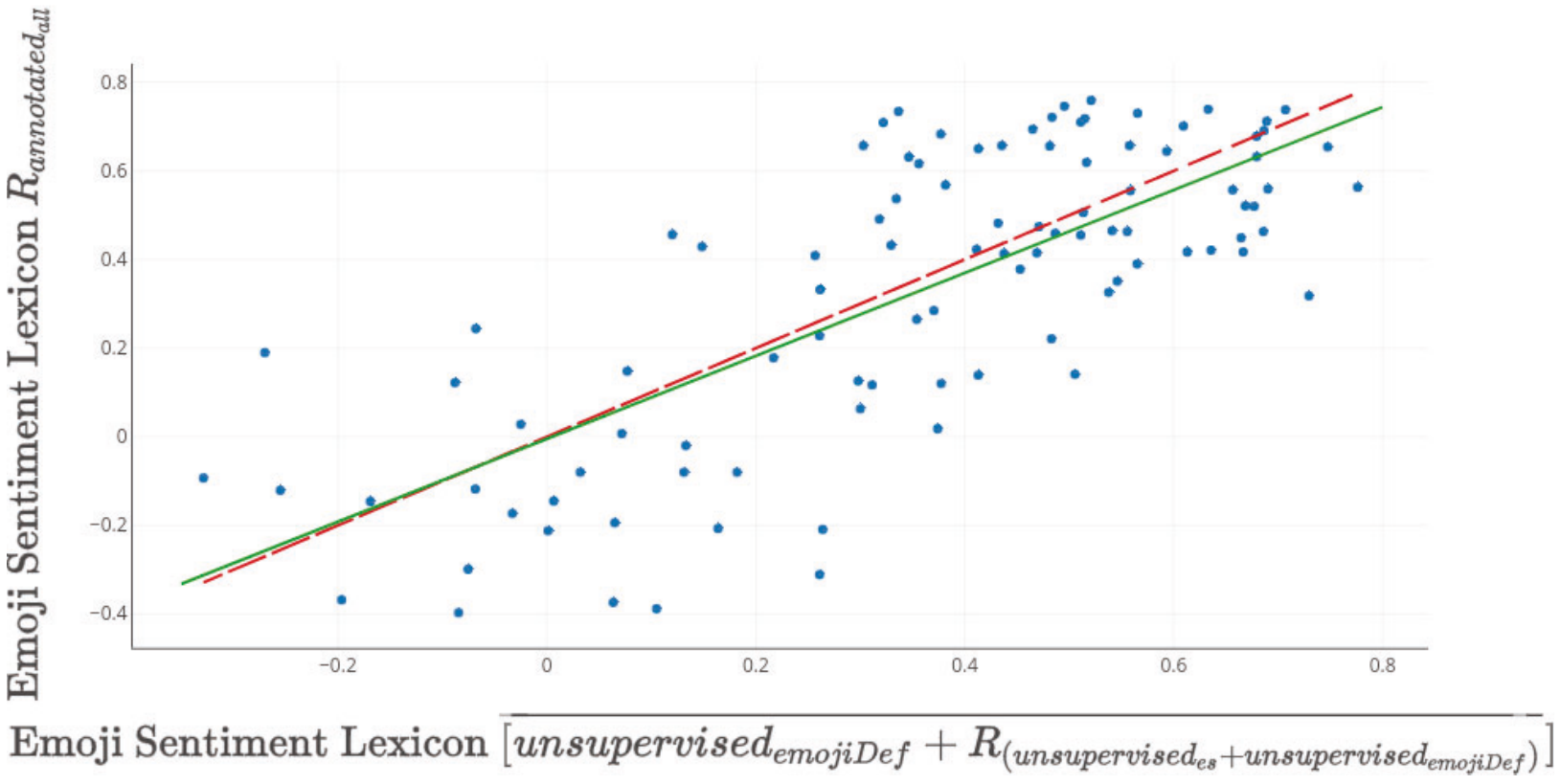

Figure 12: Scatterplot for top 100 *emoji* sentiment scores comparing [ $\overline{unsup._{emojiDef} + R_{(unsup._{es}+unsup._{emojiDef})}}$ ] with $R_{annotated_{all}}$

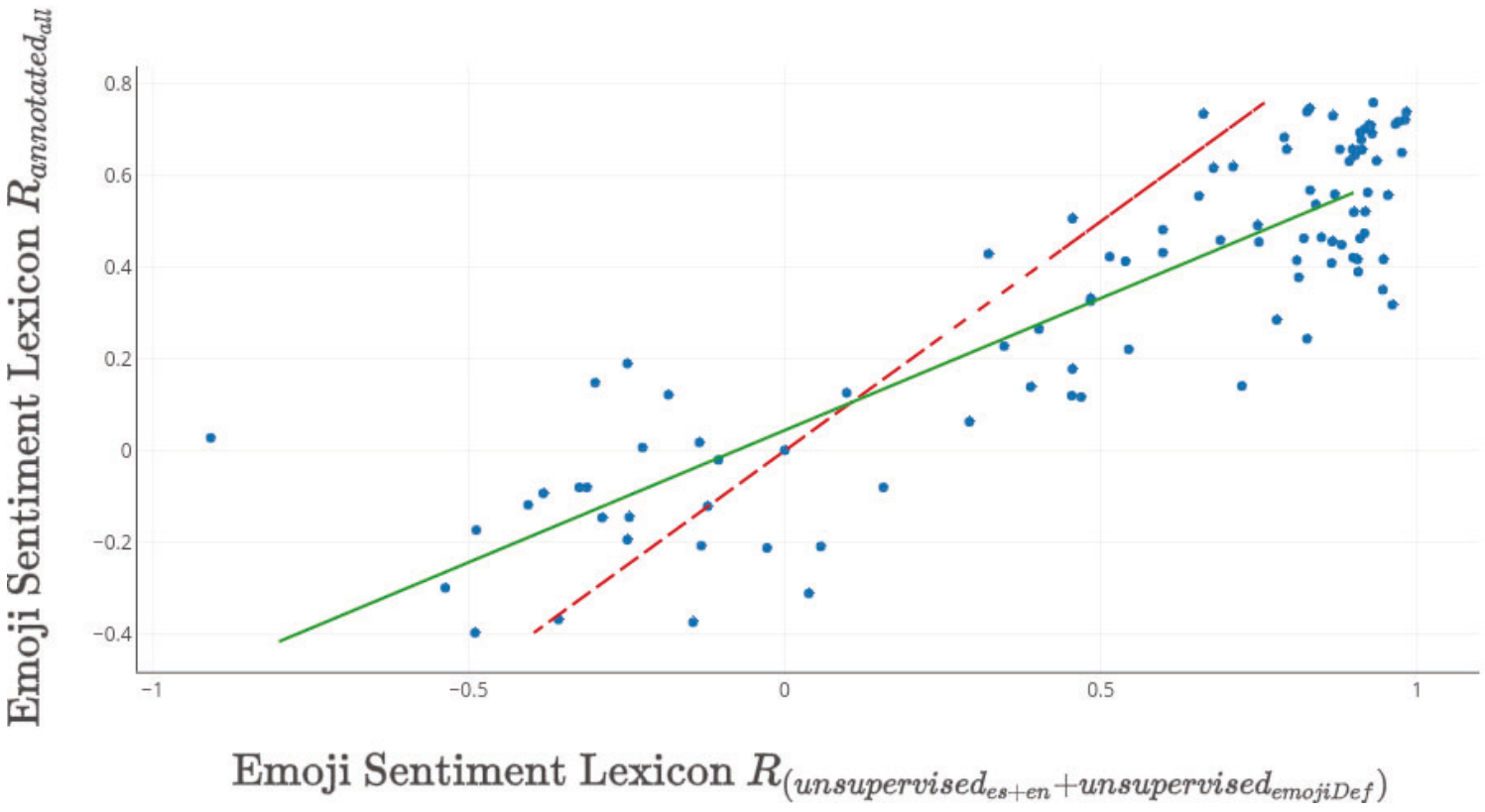

Figure 13: Scatterplot for top 100 *emoji* sentiment scores comparing $R_{(unsupervised_{es+en}+unsupervised_{emojiDef})}$ with $R_{annotated_{all}}$

Figure 10 shows the scatterplot of the sentiment scores of the top 100 *emojis* for lexica $unsupervised_{emojiDef}$ and$R_{annotated_{all}}$. We observe that these lexica are softly correlated. Note that $R_{annotated_{all}}$ is biased by the view of the creator and to a lesser extent by the vision of the user.

Figures 11, 12 and 13 show, respectively, the scatterplots when comparing the lexica obtained from the English, Spanish and Spanish+English datasets. We only provide one example per variant of our method, since all scatterplots are quite similar. We observe that, in the case of the English, the regression and gold-standard lines cross at positive *emoji* sentiments. In the case of Spanish, both lines are quite similar. For Spanish+English, the lines cross at neutral *emoji* sentiments.

We also measured the Pearson and Spearman correlations between our approach and its enhanced variants, on the one hand, and the $R_{annotated_{all}}$ lexicon, on the other. *Emoji* sentiment scores and occurrences were separately sorted in descending order, and both correlation metrics were estimated as in Eq(4):

$$r(x, y) = \frac{\sum_{i=1}^{n}(x_{e_i} - \bar{x})(y_{e_i} - \bar{y})}{\sqrt{\sum_{i=1}^{n}(x_{e_i} - \bar{x})^2 \sum_{i=1}^{n}(y_{e_i} - \bar{y})^2}}, \tag{4}$$

where $e_i$ are the *emojis*; $x_{e_i}$ and $y_{e_i}$ are the variables of our lexica and $R_{annotated_{all}}$, respectively, $i = 1, \cdots, n$; $\bar{x}$ and $\bar{y}$ are their averages; and $n$ is the number of *emojis*. For the Pearson correlation, which we denote by $r_{score}(x, y)$, we correlated the *emojis* ranked by score and occurence in $R_{annotated_{all}}$. The Spearman correlation, which we denote by $r_{rank}(x, y)$, was computed replacing $x$ and $y$ scores by ranks.

Similarly, we compared the results with the lexicon obtained by appying LIWC on *emoji*CLDR names with the results obtained by extracting sentiment words from *WordNet-Affect*. Table 7 illustrates the outcome with our lexica, for the 100 top *emoji* sentiment scores and ranks. Table 8 illustrates the case for the top 150 ones. In the first table we can observe that (E2) is the best approach for all the datasets, compared with $R_{annotated_{all}}$, but if we consider more *emojis*, the best approach is (E3) regardless of the dataset. Moreover, we can observe the poor results with LIWC. Even though the approach with *WordNet-Affect* is very effective for the very top *emojis*, if we add more *emojis* its Spearman correlation decreases.

| **Lexicon** $x$ | **Lexicon** $y$ | | | $r_{score}(x, y)$ | $r_{rank}(x, y)$ |
|---|---|---|---|---|---|
| $R_{annotated_{all}}$ | **Our approach** | | $unsupervised_{emojiDef}$ | 68.42% | 68.65% |
| | | (E1) | $R_{unsupervised_{es}}$ | 74.93% | 74.78% |
| | | | $R_{unsupervised_{en}}$ | 79.70% | 75.25% |
| | | | $R_{unsupervised_{es+en}}$ | 84.21% | 77.88% |
| | | (E2) | $R_{(unsupervised_{es}+unsupervised_{emojiDef})}$ | 81.32% | 79.07% |
| | | | $R_{(unsupervised_{en}+unsupervised_{emojiDef})}$ | 86.90% | 80.71% |
| | | | $R_{(unsupervised_{es+en}+unsupervised_{emojiDef})}$ | 87.39% | 80.75% |
| | | (E3) | $[\ \overline{unsup._{emojiDef} + R_{(unsup._{es}+unsup._{emojiDef})}}\ ]$ | 74.66% | 69.86% |
| | | | $[\ \overline{unsup._{emojiDef} + R_{(unsup._{en}+unsup._{emojiDef})}}\ ]$ | 83.70% | 78.84% |
| | | | $[\ \overline{unsup._{emojiDef} + R_{(unsup._{es+en}+unsup._{emojiDef})}}\ ]$ | 83.16% | 78.06% |
| | **Lu et al., 2016** | | LIWC$_{emoji\text{CLDR}}$ | 39.88% | 41.93% |
| | **Kimura et al., 2017** | | WordNet-Affect$_{emoji}$ | 89.10% | 89.62% |

Table 7: Score and rank correlations considering the 100 *emojis* with more occurrences ranked by score and occurrences.

## 5. Conclusions

In this paper we describe an unsupervised SA strategy based on semantic dependencies, called USSPAD, enhanced with SA of descriptions by *emoji* creators from `Emojipedia`, with the objective of creating a fully unsupervised *emoji* sentiment lexicon. This lexicon is then improved in different variants that take advantage of the sentiment distribution of informal texts including *emojis*. In all cases, USSPAD guarantees that neither labeling nor training is necessary. Our approach and its variants are applied to the Spanish, English and Spanish+English datasets provided in (Kralj Novak et al., 2015) and demonstrate good performance in comparison with the lexica of (Kralj Novak et al., 2015) and (Lu et al., 2016).

| Lexicon $x$ | Lexicon $y$ | | $r_{score}(x,y)$ | $r_{rank}(x,y)$ |
|---|---|---|---|---|
| $R_{annotated_{all}}$ | **Our approach** | $unsupervised_{emojiDef}$ | 57.17% | 58.75% |
| | | (E1) $R_{unsupervised_{es}}$ | 60.95% | 60.97% |
| | | (E1) $R_{unsupervised_{en}}$ | 72.15% | 66.91% |
| | | (E1) $R_{unsupervised_{es+en}}$ | 77.54% | 69.52% |
| | | (E2) $R_{(unsupervised_{es}+unsupervised_{emojiDef})}$ | 75.02% | 74.96% |
| | | (E2) $R_{(unsupervised_{en}+unsupervised_{emojiDef})}$ | 77.67% | 74.41% |
| | | (E2) $R_{(unsupervised_{es+en}+unsupervised_{emojiDef})}$ | 81.90% | 79.34% |
| | | (E3) $[\ \overline{unsup._{emojiDef} + R_{(unsup._{es}+unsup._{emojiDef})}}\ ]$ | 70.17% | 69.03% |
| | | (E3) $[\ \overline{unsup._{emojiDef} + R_{(unsup._{en}+unsup._{emojiDef})}}\ ]$ | 73.77% | 70.42% |
| | | (E3) $[\ \overline{unsup._{emojiDef} + R_{(unsup._{es+en}+unsup._{emojiDef})}}\ ]$ | 81.90% | 73.04% |
| | **Kimura et al., 2017** | WordNet-Affect$_{emoji}$ | 87.73% | 52.03% |

Table 8: Score and rank correlations considering the 150 *emojis* with more occurrences ranked by score and occurrences.

Our approach analyzes dependencies between lemmatized tagged words using a sentiment propagation algorithm that considers key linguistic phenomena, namely intensification, modification, negation, and adversative and concessive relations. All of these, but specially negation, play an important role in SA. Our approach and its variants are sophisticated (and therefore more precise) than the simplistic unsupervised approach of (Lu et al., 2016), which only takes into account the polarity of isolated words using LIWC.

Experiments using various datasets in different languages demonstrate quite satisfactory results for the basic approach and for variants, in comparison with results using *emoji* sentiment lexica acquired from short CLDR names and manually labelled by several human annotators. We obtain improvements of 2%-8% in terms of accuracy, 6%-25% in terms of $P_{macro}$ and 2%-11% in terms of $F_{macro}$ depending on the dataset language and the variant. Comparing them with the manually annotated data provided by (Kralj Novak et al., 2015), in most cases we are only between 1% and 3% worse in terms of accuracy, $P_{macro}$ and $F_{macro}$, even though in some cases we achieve better results.

To the best of our knowledge, this is the first time that definitions from `Emojipedia` have been applied in an approach to automatically determine the sentiment polarities of *emojis*, using for this purpose an USSPAD. It is also the first time that different variants of *emoji* sentiment lexica have been created from that information, taking advantage of the messages they are embedded in.

Even though supervised algorithms should be expected to attain better performance rates, they have the drawback that they require specifically annotated corpora for each domain of interest, which in turn requires human effort for the labeling of documents involved in training. In our case, no training or labeled datasets are required. The latter are often difficult or even impossible to obtain (Moreno Ortiz & Pérez Hernández, 2013), as their generation is too labor-intensive and time-consuming.

As our results are promising, we envisage several directions for future work. First, we would like to verify our findings on other collections of texts. We would then like to expand the meaning definitions and use repositories other than `Emojipedia`. Last, we would like to exploit structural and semantic aspects of text in order to identify important and less important text spans in *emoji*-based sentiment analysis.

## Acknowledgments

We wish to thank Mayo Kimura and Marie Katsurai for providing us with their *emoji* sentiment lexicon in (Kimura & Katsurai, 2017). This work was partially supported by Mineco grant TEC2016-76465-C2-2-R and by Xunta de Galicia grant GRC2014/046, Spain.